%% file: main.tex
\documentclass[11pt,letterpaper]{article}
\usepackage{livecraft}
\usepackage{array,tabularx,longtable}
\usepackage{multirow}
\usepackage{placeins}
\usepackage{flafter}
\usepackage{capt-of}
\usepackage{microtype}
\usepackage[pagebackref,breaklinks,colorlinks,linkcolor=LCBlueText,citecolor=LCBlueText,urlcolor=LCGreenText]{hyperref}
\newlength{\designwidth}\AtBeginDocument{\setlength{\designwidth}{\textwidth}}
\newcommand{\designscale}{1.2547}
\newcommand{\benchmark}{LiveCraft}
\newcommand{\supp}{appendix}
\title{\textcolor{LCBrand}{Live}Craft: Benchmarking Long-Horizon Visual Agents\\in Real-Time Open Worlds}
\author{Chengxing Xie\textsuperscript{1}, Deyao Hong\textsuperscript{1}, Erle Zhu\textsuperscript{1}, Mingyu Su\textsuperscript{1}, Tongkai Yang\textsuperscript{2}, Hongning Wang\textsuperscript{1}}
\lcaffil{\lcinst{tsinghua}{1}{The Conversational AI (CoAI) Group, Tsinghua University}\hspace{1.6em}\lcinst{antgroup}{2}{Ant Group}\\[0.8mm]
Correspondence to: \mbox{Chengxing Xie \textless\href{mailto:xiechengxing34@gmail.com}{xiechengxing34@gmail.com}\textgreater}, \mbox{Hongning Wang \textless\href{mailto:hwai@tsinghua.edu.cn}{hwai@tsinghua.edu.cn}\textgreater}}
\hypersetup{pdfauthor={Chengxing Xie, Deyao Hong, Erle Zhu, Mingyu Su, Tongkai Yang, Hongning Wang}}
\newcommand{\lcurlsite}{https://livecraft-benchmark.github.io/}
\newcommand{\lcurlcode}{https://github.com/livecraft-benchmark/livecraft}
\newcommand{\lcurldata}{https://huggingface.co/datasets/livecraft-benchmark/livecraft-maps}
\lclinks{\lclink[\lcurlsite]{\faGlobe}{Project Page}\hspace{.5em}\lclink[\lcurlsite\#leaderboard]{\faTrophy}{Leaderboard}\hspace{.5em}\lclink[\lcurlcode]{\faGithub}{Code}\hspace{.5em}\lclink[\lcurldata]{\faDatabase}{Dataset}}
\begin{document}
\brokenpenalty=10000
\maketitle
\begin{center}
\includegraphics[width=\textwidth]{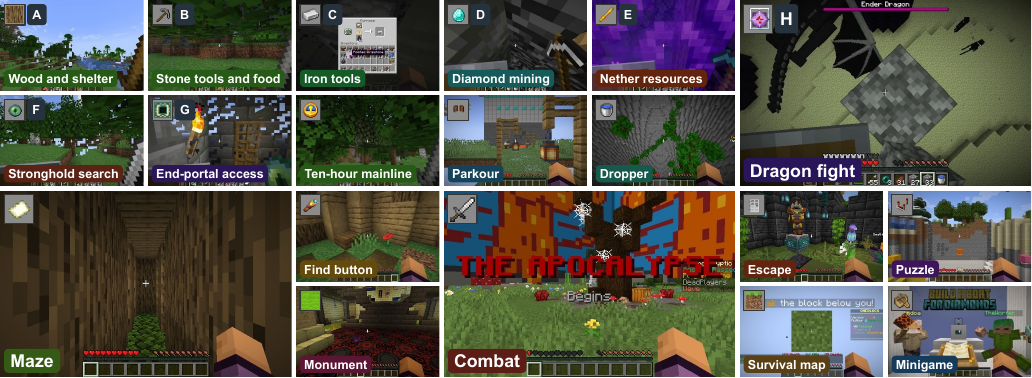}
\vspace{-1mm}\captionof{figure}{Gameplay in \benchmark{}: eight survival stages, a ten-hour mainline and one map from each of ten categories.}
\label{fig:title-mosaic}
\end{center}
\input{sections/abstract}
\input{sections/introduction}
\input{sections/related_work}
\input{sections/benchmark}
\input{sections/experiments}

\input{sections/conclusion}
{
\small
\bibliographystyle{ieeenat_fullname}
\bibliography{references}
}
\clearpage
\lcbanner{Appendix}
\vspace{.5em}
\appendix
\setcounter{topnumber}{4}\setcounter{bottomnumber}{4}\setcounter{totalnumber}{6}
\makeatletter\setlength{\@fptop}{0pt}\makeatother
\input{sections/appendix_results}
\input{sections/appendix_dataset}
\input{sections/appendix_tasks}
\input{sections/appendix_gallery}
\input{sections/appendix_tools}
\input{sections/appendix_limitations}

\input{sections/statements}
\input{sections/appendix_maps}
\end{document}

%% file: sections/abstract.tex
\begin{abstract}
Large language model (LLM) agents often work in stable workspaces that remain unchanged while they think. However, many real-world tasks require real-time perception and interaction with dynamic environments. To evaluate this capability, we introduce \benchmark{}, a benchmark for long-horizon visual agents in real-time open worlds. Built on the open-world game Minecraft, \benchmark{} contains eight survival stages, a continuous mainline, and 70 community maps. The default task set, \benchmark{}-Lite, contains the stages, the mainline and 20 Lite maps, and \benchmark{}-Full adds the other 50 maps. The survival tasks follow the progression toward defeating the Ender Dragon, with checkpoints from built-in advancements and task-specific goals. Community maps add parkour, mazes, dropper challenges and puzzles with different task demands on reaction speed and memory. In the primary Screen real-time (Screen-RT) setting, agents access the game only through screenshots and keyboard--mouse actions while the world keeps running, and a separate verifier checks progress directly from the game's world state. We evaluate four frontier agent systems built on Claude Code and Codex. Under Screen-RT, Claude Opus 5.5 with Claude Code achieves the highest mean scores, 30.3 on the stages and 10.8 on the Lite maps out of 100. Pausing the world tends to lower the death rate per hour of game time. Basic tools, which replace screenshots and keyboard--mouse control with tool calls, raise real-time stage scores to 1.7--2.7$\times$ those of Screen-RT. Together, these results show that current agent systems still have substantial room for improvement in long-horizon visual interaction with real-time open worlds.
\end{abstract}

%% file: sections/introduction.tex
\clearpage
\input{sections/teaser}
\section{Introduction}
\label{sec:introduction}
Large language model (LLM) agents now solve many complex software engineering tasks through code execution and tool feedback. On Terminal-Bench 4.0~\citep{terminalbench4leaderboard2026}, Claude Opus 5.5 with Claude Code achieves a 64.8\% task-resolution rate at maximum reasoning effort. In controlled software environments, the environment state can remain unchanged during reasoning, while compiler messages, test results and command outputs provide structured feedback for revising actions. Agents already perform well in such settings.

Recent work is extending these agents toward physical interactions: Project Fetch~\citep{projectfetch2026}, for example, uses Claude Code to operate robot sensing and control programs, and vision-language-action models such as RT-2~\citep{rt22023} and OpenVLA~\citep{openvla2024} map camera input to robot control. Benchmarks such as MineExplorer~\citep{mineexplorer2026} and GameWorld~\citep{gameworld2026} evaluate agents in games, yet their main protocols do not jointly test multi-hour goals, native visual control and a world that continues evolving during model reasoning. \textbf{The open question is whether agents can complete long-horizon tasks in a dynamic open world using visual observations and basic actions, without pausing the world or relying on structured state and action interfaces.}

To address this gap, we introduce \benchmark{}, a benchmark for long-horizon visual agents in real-time open worlds, built on the open-world game Minecraft. In our primary setting, Screen real-time (Screen-RT), an agent receives a concrete goal and interacts only through screenshots and keyboard--mouse actions while the world keeps running. For example, in Figure~\ref{fig:teaser} the goal is to prepare stone equipment, lighting and food for underground exploration. The agent crafts tools, a furnace and torches and descends underground, but does not finish with the required supplies. Alongside Screen-RT, we compare a turn-based clock and Basic and Macro tools with local or high-level actions.

Different tasks demand different capabilities, such as fast reaction or long-horizon memory. To cover these task demands, we collect tasks from two sources. First, we organize the survival progression toward defeating the Ender Dragon into eight separately initialized stages and a continuous mainline that starts empty-handed. Second, we collect community maps across ten mechanic categories, including mazes, parkour and puzzles, spanning both rapid reactive control and long-horizon memory. The benchmark has two task sets. The default set, \textbf{\benchmark{}-Lite}, contains the eight stages, the continuous mainline and 20 Lite maps, for 29 tasks, and \textbf{\benchmark{}-Full} adds the remaining 50 maps, for 79 tasks. The Lite maps comprise two per category with full-score walkthrough videos for quality checks. Figures~\ref{fig:overview} and~\ref{fig:composition} illustrate the interaction process and task composition.

To measure how far an agent progresses even when it cannot finish, we build a separate verifier that tests explicit conditions directly on the game's world state, such as the player's inventory, position and advancements, which the agent cannot see. Each check links credit to a defined task outcome, such as obtaining an item, reaching a location or advancing a map counter. Fine-grained checkpoints distinguish agents that reach different parts of a task: survival checks follow documented advancement descriptions and task-specific goals, while community checks use map-native level counters, scoreboards and other progress signals. The verifier records when each checkpoint is first satisfied, allowing partial trajectories to be compared as well as final success. These code-based tests make the scored outcomes explicit while the evaluated agent still interacts visually.

We evaluate four frontier agent systems built on Claude Code and Codex. \textbf{Under the primary Screen-RT setting, no evaluated system achieves full credit on any \benchmark{}-Lite task.} Claude Opus 5.5 with Claude Code achieves the highest mean task scores, 30.3 on the stages and 10.8 on the Lite maps, far from full credit. Pausing the world tends to lower the death rate per hour of game time but does not let any system solve a stage with the Screen interface. Relieving time pressure is thus not enough, and our ablations point to visual perception, low-level control and long-horizon memory as the bottlenecks. High- and low-frequency maps also behave very differently: Basic tools nearly triple scores on low-frequency maps, whereas every real-time setting stays at or below 5 points on high-frequency maps, which shows the limit of the evaluated systems when a task demands fast decisions. These findings motivate improvements in visual perception and understanding, low-level control, long-horizon memory and adaptation to unfamiliar task mechanics, and faster responses to a running world.

%% file: sections/teaser.tex
\begin{figure*}[h!]
\centering
\includegraphics[width=\textwidth]{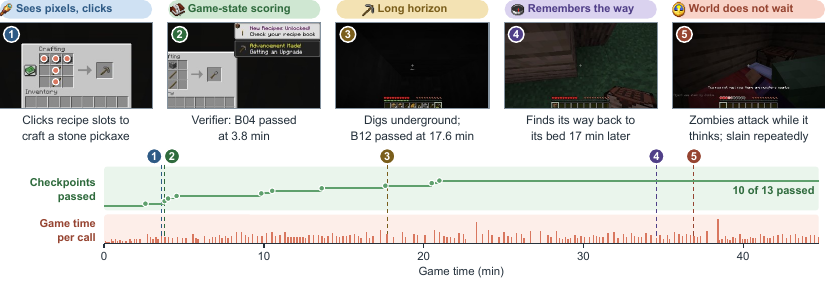}
\caption{One recorded Screen-RT run of Claude Opus 5.5 on one survival stage. The agent acts only by clicking and pressing keys on what it sees, while a separate verifier records checkpoints from the game's world state. The world keeps running while the agent thinks, so zombies attack between calls and the run ends at the death limit. Numbers mark each frame on the timeline, which shows the checkpoints passed and the game time that elapsed during each call.}
\label{fig:teaser}
\end{figure*}

%% file: sections/related_work.tex
\section{Related Work}
\label{sec:related-work}

\paragraph{Visual and embodied agents.}
Interactive benchmarks evaluate agent control across web, desktop and mobile interfaces in WebArena~\citep{webarena2024}, VisualWebArena~\citep{visualwebarena2024}, OSWorld~\citep{osworld2024} and AndroidWorld~\citep{androidworld2025}, as well as multi-stage embodied manipulation in ALFRED~\citep{alfred2020}, CALVIN~\citep{calvin2022} and LIBERO~\citep{libero2023}. Some benchmarks~\citep{longpuzzlebench2026,visualcipherbench2026} test long-horizon visual reasoning on puzzles. Game environments have similarly progressed from the Arcade Learning Environment~\citep{arcadelearning2013} and real-time strategy in AlphaStar~\citep{alphastar2019} to language-guided 3D gameplay in SIMA~\citep{sima2024}. Recent suites broaden game coverage, from retro games played from raw frames to multi-game arenas and browser games with state-checked outcomes and real-time variants~\citep{videogamebench2025,omnigamearena2026,gameworld2026}. These suites cover many game mechanics, but none places agents in a persistent, real-time open world with such a wide range of objectives. \benchmark{} brings these dimensions together in a world that agents control through screenshots, keyboard and mouse.

\paragraph{Minecraft agents.}
Within this broader setting, Minecraft links resource gathering, crafting and exploration into long chains of interdependent actions, established as a sequential decision-making testbed by MineRL~\citep{minerl2019} and MineDojo~\citep{minedojo2022} and studied through pixel-based imitation in VPT~\citep{vpt2022}. Recent benchmarks evaluate visual agents on multi-step objectives: MCU~\citep{mcu2025} scores composable tasks with vision-language criteria, while MineExplorer~\citep{mineexplorer2026} tests prerequisite chains with milestone checks under a fixed simulation-step budget. Meanwhile, agent frameworks pursue extended progression: Voyager~\citep{voyager2023} and GITM~\citep{gitm2023} pair explicit state with executable skills and structured actions, while JARVIS-1~\citep{jarvis12023} plans over visual observations with multimodal memory. These works establish multi-step Minecraft capabilities, but their evaluations do not jointly cover multi-hour survival progression and diverse community mechanics under an uninterrupted visual interaction loop. \benchmark{} evaluates long-horizon goals across resource gathering, exploration, parkour, mazes, combat and puzzles, with a state-based verifier recording progress while the world keeps running.

To connect these lines of work, \benchmark{} unites native visual interaction, real-time world dynamics, long-horizon goals and diverse task mechanics in a single evaluation, paired with a state-based verifier and controlled clock and interface variants for diagnosis.

%% file: sections/benchmark.tex
\section{\benchmark{}}
\label{sec:benchmark}
\begin{figure*}[t]
\centering\begin{minipage}{\designwidth}
\centering
\includegraphics[width=\linewidth]{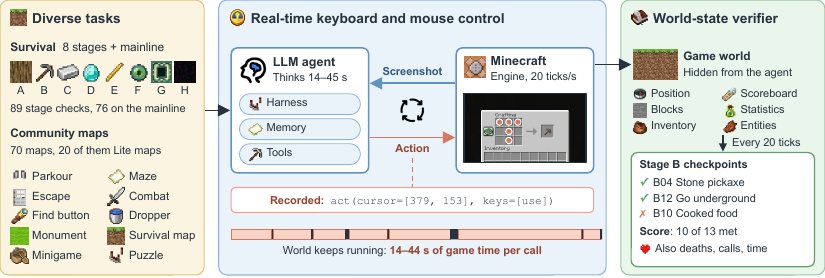}
\caption{Overview of \benchmark{}. An LLM agent plays diverse tasks by viewing screenshots and controlling a running Minecraft engine with keyboard and mouse. A separate verifier scores progress from the game's world state, which is hidden from the agent.}
\label{fig:overview}
\end{minipage}
\end{figure*}
\benchmark{} tests whether LLM agents can complete diverse visual tasks with different task demands, such as fast reaction and long-horizon memory, while the world keeps running. Built on MineStudio~\citep{minestudio2025}, it contains 79 tasks: eight initialized survival stages, one continuous mainline and 70 community maps. As shown in Figure~\ref{fig:overview}, each run connects a task, the agent's interaction loop and a separate verifier. In the primary Screen-RT setting, the agent receives a goal, observes screenshots and acts through keyboard--mouse actions, while the verifier reads the game's world state to record checkpoints. This setup evaluates how agents observe, act and sustain progress in a running world, with turn-based and tool variants testing which difficulties remain when real-time pressure is removed or interface support is added.

\subsection{Task Design}
\label{sec:tasks}
\begin{figure*}[t]
\centering\begin{minipage}{\designwidth}
\centering
\begin{subfigure}[b]{0.445\linewidth}
\centering
\includegraphics[width=\linewidth]{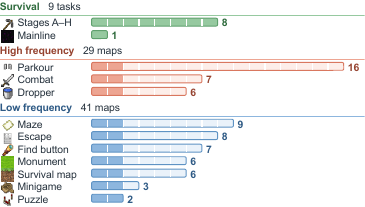}
\caption{Tasks by Source and Category}
\label{fig:composition-tasks}
\end{subfigure}\hfill
\begin{subfigure}[b]{0.535\linewidth}
\centering
\includegraphics[width=\linewidth]{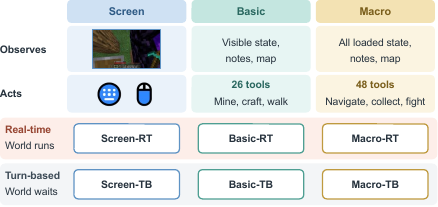}
\caption{Interaction Settings}
\label{fig:composition-settings}
\end{subfigure}
\caption{Tasks and interaction settings. (a) Tasks grouped by source and by whether they demand high- or low-frequency decisions; each bar has one segment per task, and dark segments mark the tasks of \benchmark{}-Lite. (b) Each setting combines an interface, which fixes what the agent observes and how it acts, with a world clock.}
\label{fig:composition}
\end{minipage}
\end{figure*}

To test both sustained progress toward a goal and adaptation to different task demands, we combine Minecraft survival tasks with community-authored challenges. Survival stages make individual milestones accessible, the continuous mainline tests whether an agent can establish and carry forward their prerequisites, and community maps add objectives that demand fast reaction or long-horizon memory.

\paragraph{Survival stages.}
We first use Minecraft's survival progression toward defeating the Ender Dragon, which links resource gathering, crafting and exploration. Evaluating only the complete progression makes later capabilities difficult to assess when an agent fails early. Minecraft's built-in advancements~\citep{minecraftadvancements2017} provide a guide through the game. They include entering new dimensions and defeating the dragon. We use these advancement conditions together with task-specific goals to organize eight initialized stages, A to H, as shown in the left part of Figure~\ref{fig:overview}. Each stage starts from a prepared world state on a fixed seed, so the agent can attempt its objective independently. Stage E, for example, supplies a lit Nether portal and equipment, then requires the agent to obtain blaze rods and ender pearls and return to the Overworld. Details of the task collections and checkpoints are provided in the \supp{}.

\paragraph{Continuous mainline.}
To test whether agents can connect these milestones, we also define a continuous mainline: start empty-handed and defeat the Ender Dragon within ten hours. Resources, world changes and accumulated information carry forward, so later objectives depend on earlier actions. To measure partial progress along this long trajectory, we define 76 scored checkpoints derived from the stage checks. Unlike the initialized stages, which supply the resources and world state needed to attempt a particular objective, the continuous mainline starts empty-handed and requires the agent to establish those prerequisites itself, so we report its score in a separate column of Table~\ref{tab:main-results}.

\paragraph{Community maps.}
Survival's resource-and-crafting progression covers a limited range of scenarios and task demands. To address this limitation, we collect community maps that broaden both the situations agents encounter and the pace of interaction. We screen maps for \emph{mechanic coverage, route and difficulty variation, version compatibility and observable progress signals}, selecting 70 maps. As shown in Figure~\ref{fig:composition-tasks}, they cover ten mechanic categories: parkour, combat, dropper challenges, mazes, puzzles, and others. \textbf{Combat requires rapid responses to threats, while mazes require remembering routes over many observations.} Each map keeps its own start and goal. To reduce the cost of evaluating the full collection while retaining category coverage, we select two maps per category as the 20 Lite maps. The default task set, \textbf{\benchmark{}-Lite}, contains the eight survival stages, the continuous mainline and the Lite maps, 29 tasks in total. \textbf{\benchmark{}-Full} adds the other 50 maps, for 79 tasks. For quality review, each of the 20 Lite maps has a full-score walkthrough video. We adapted the selected maps to our Minecraft 1.21.1 engine and checked that each starts.

\subsection{Interaction and World Clocks}
\label{sec:interfaces}
\begin{figure*}[t]
\centering
\includegraphics[width=\textwidth]{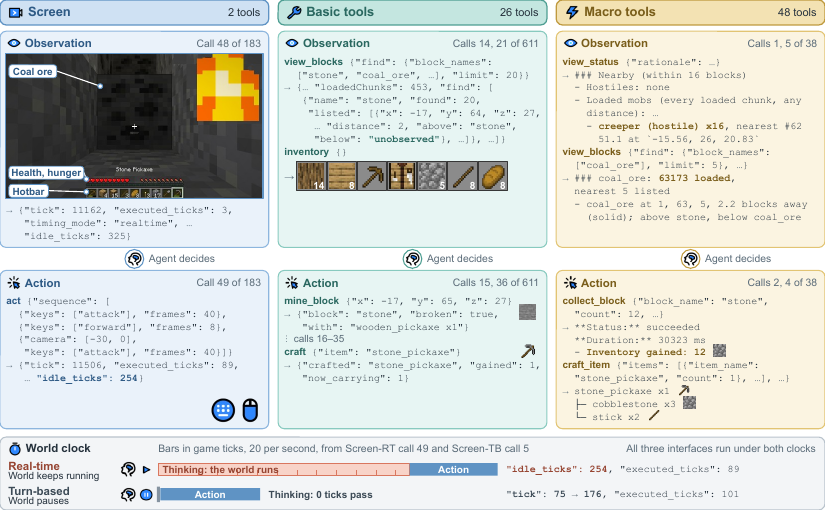}
\caption{What the agent observes and how it acts under each interface, from recorded calls of Claude Opus 5.5 on survival stage B. Each column shows observations and actions from one run, numbered by call, with tool inputs and responses shortened by ellipses. The bottom band shows the game ticks of one recorded call under each world clock: the world keeps running while the agent thinks under the real-time clock and pauses under the turn-based clock.}
\label{fig:setting-views}
\end{figure*}
With the task environments fixed, we next define how agents observe and act within them. Our primary setting uses screenshots and keyboard--mouse control in a continuously running world. \textbf{As in real-time robotic control, the agent must interpret visual observations and choose actions while the environment continues to evolve.} To test whether failures arise from interpreting screenshots, executing actions or responding before the world changes, we vary the interface and world clock. As illustrated in Figure~\ref{fig:setting-views}, tools provide explicit state and automate actions to different degrees, and turn-based settings pause the world during reasoning.

\paragraph{Screen interaction.}
To evaluate direct visual control, Screen-RT exposes the screen and keyboard--mouse interface shown in Figure~\ref{fig:composition-settings}. The \texttt{look} tool returns a $960\times540$ screenshot, and \texttt{act} holds keys, moves the camera or clicks for up to 400 game ticks before returning another screenshot. Between calls, keys are released and no new frames arrive, while the world advances during thinking and request processing at a nominal 20 ticks per second. Tool responses report action-execution and between-call tick counts, allowing the model to track elapsed game time.

\paragraph{Clock variants.}
To test the effect of world evolution during reasoning, Screen turn-based (Screen-TB) pauses the world between calls and advances it during action execution, so thinking consumes wall-clock budget without advancing game time. Slowed variants instead retain continuous evolution at 0.5 or 0.25 times normal speed. All preserve the screen and keyboard--mouse interface.

\paragraph{Information and action support.}
To examine whether explicit state information and tool-based actions, which replace keyboard--mouse control with structured tool calls, help agents sustain progress, we provide additional interface support. Basic tools are designed to approximate the local actions available to an ordinary player, such as moving, mining a block and crafting a recipe, while exposing them as structured calls. They replace screenshots with basic state queries and provide a remembered map and notes.

To test agents with more extensive automation, Macro tools provide high-level action routines and privileged state access. Our Macro tools, adapted from mine-ai-mcp~\citep{mineaimcp2026}, expose loaded-chunk information even for occluded objects and automate pathfinding, batch collection, recursive crafting, survival responses and dragon combat, while keeping the local actions of the Basic tools. This setting tests how far agents can progress when the interface supplies information and control support beyond ordinary screen interaction. Both tool sets run under both clocks, and the \supp{} lists every tool.

\subsection{Verifier Construction and Scoring}
\label{sec:scoring}
To measure partial progress and completion, a separate verifier reads the game's world state directly, such as the player's inventory, position and advancements, and records when each task-defined checkpoint is satisfied.

\paragraph{Progress scoring.}
Survival tasks combine built-in advancements, such as entering the Nether or defeating the Ender Dragon, with finer-grained checks for collecting materials and crafting tools. Survival scores are the percentage of checkpoints satisfied. Community maps define weighted units from native mechanics and progress signals totaling 100 points per map: aMAZEing uses twelve position and transition units across three mazes, and Apocalyptic awards five points per wave reached, up to twenty waves before the first death. Collection means weight tasks equally rather than pooling checkpoints.

\paragraph{Deaths and survival.}
We also record deaths as a signal of agent capability. A death returns the player to the spawn point, so the agent must find its way back before it can continue. Recorded checkpoints stay credited, so deaths do not lower the score; they show how well an agent stays alive while pursuing the goal.

%% file: sections/experiments.tex
\section{Experiments}
\label{sec:experiments}
\input{sections/setup}
\input{sections/results}

%% file: sections/setup.tex
\subsection{Evaluation Setting}
\label{sec:setup}
The main evaluation tests four agent systems on \benchmark{}-Lite across six settings that pair the Screen interface, Basic tools and Macro tools with the real-time and turn-based clocks. Each task--setting pair has one run on a fixed world seed, with a wall-clock budget of two hours per survival stage, two or three hours per community map and ten hours for the continuous mainline.

\paragraph{Agent systems.}
We evaluate \ico{deepseek}\,DeepSeek V4.1 Flash~\citep{deepseekv41flash2026}, \ico{glm}\,GLM-5.3-flash~\citep{glm53flash2026} and \ico{claude}\,Claude Opus 5.5~\citep{claudeopus55_2026} in Claude Code 2.1.306, and \ico{openai}\,GPT-6-astra~\citep{gpt6astra2026} in Codex 0.165.0, all at medium reasoning effort. To standardize context budgets and bound the size of image-heavy requests, we give every system the same 200K-token context window.

\paragraph{Ablations.}
To find which capability limits the agents, we add ablations on the world clock, the observation and the action. For the world clock, we slow the world to half and quarter speed and compare the outcome with the paused world (Section~\ref{sec:clock-results}). For the observation, we add the player's state as text to each screenshot, or replace the screenshots with text observation (Section~\ref{sec:observation-results}). For the action, Basic-tool actions replace keyboard--mouse control (Section~\ref{sec:action-results}).

\paragraph{Metrics.}
The benchmark score of a task set is the mean task score over all of its tasks, 29 for \benchmark{}-Lite and 79 for \benchmark{}-Full. For each task collection, we also report the mean task score, the number of solved runs and the mean deaths and model turns per run. A run solves a task only when it earns full credit, with every checkpoint satisfied. A run ends when the task is solved, the agent stops, a task-specific failure occurs, or the time or death limit is reached; the death limit is 200 on the mainline and 50 elsewhere. Per-system stage scores, run endings, timing and case studies are in the \supp{}.

%% file: sections/results.tex
\subsection{Scores under Different Interaction Settings}
\label{sec:overall}
\label{sec:results}

\input{sections/table_main}
We first evaluate the four systems in the primary Screen-RT setting, where agents act through screenshots and keyboard--mouse actions while the world keeps running. \textbf{Under Screen-RT, no system solves any of the 29 \benchmark{}-Lite tasks.} \benchmark{}-Lite scores range from 7.3 to 17.0, as reported in Table~\ref{tab:main-results}.

The four systems fail in different ways. Claude Opus 5.5 is the strongest: it has the highest \benchmark{}-Lite score in every setting, and under Screen-RT it leads on the stages (30.3), the Lite maps (10.8) and the continuous mainline (32.9). DeepSeek V4.1 Flash nearly matches it on the stages (27.7) but dies most often, 27.0 times per stage run, and scores lowest on the Lite maps (5.9). GLM-5.3-flash is weaker on the stages (12.8) and second on the Lite maps (7.4). GPT-6-astra has the lowest \benchmark{}-Lite score together with the fewest deaths and turns, because it ends its runs early, as analyzed below. We also run all four systems under Screen-RT on \benchmark{}-Full and report these results in the \supp{}: Claude Opus 5.5 is again the strongest and the only system that solves any task, three community maps outside the Lite set.

\input{sections/results_figure}
Agents fail where a task requires navigation and combat rather than local actions. Under Screen-RT, credit concentrates in early stages, as shown in Figure~\ref{fig:results}c: all four systems earn partial credit on stages A (wood and shelter) and B (stone tools and food), where DeepSeek V4.1 Flash and Claude Opus 5.5 reach 76.9 on stage B. Scores drop once tasks require underground exploration and combat: no system exceeds 20.0 on stages C (iron tools), D (diamond mining) and E (Nether resources), and none earns credit on stages G (End-portal access) and H (dragon fight). On stage F (stronghold search), three systems reach 50.0 by crafting and throwing Eyes of Ender, both local actions, and none reaches the stronghold. On the continuous mainline, Screen-RT scores range from 1.3 to 32.9. Claude Opus 5.5 under Macro-TB is the only mainline run in any setting that defeats the Ender Dragon; it scores 96.1 because three other checkpoints remain unmet, so it is not counted as solved. The systems thus manage the opening of survival play, but overall performance is poor. \textbf{It is poorest where tasks demand hard control and long exploration.}

Changing the interaction interface changes performance substantially. Basic tools, which replace screenshots and keyboard--mouse control with tool calls, raise real-time stage means to \textbf{1.7--2.7$\times$} and Lite-map means to \textbf{2.0--3.4$\times$} those of Screen-RT, as shown in Figure~\ref{fig:results}a--b. Macro tools add routines that automate the survival progression. For Claude Opus 5.5, the real-time stage mean rises from 30.3 under Screen-RT to 51.2 with Basic tools and 98.9 with Macro tools. \textbf{This gain is far smaller on the community maps.} Its Lite-map mean rises from 10.8 to 28.4 with Basic tools and only to 39.6 with Macro tools, and the other three systems gain less than 4 points from the routines. The routines encode survival play and say little about parkour, mazes or puzzles. This contrast shows the diversity of the benchmark's tasks and why general control matters. \textbf{Automation built for one family of tasks loses most of its advantage on another, so task-specific tools alone do not make a general agent.}

One result surprised us: GPT-6-astra, a model with strong multimodal ability, scores lowest on \benchmark{}-Lite under Screen-RT. \textbf{Its runs show a distinct behavior: the model judges that it cannot finish a task and stops.} It ends all 16 of its Screen stage runs itself while most of the budget remains, and its final messages state that the objective is incomplete. Under Screen-RT it uses about 71 model turns per stage run, compared with 145--188 for the other systems. \textbf{Other evaluations report the same behavior: agents leave most of their time budget unused~\citep{frontiercs2budget2026}, and the median GPT-6-astra run lasts 33 minutes of a six-hour limit~\citep{frontierbioml2026}.} The effect on its scores is visible on the continuous mainline: under Basic-RT it stops after half an hour with 7.9 points, whereas under Basic-TB it uses the full ten hours and reaches 43.4, within the 39.5--59.2 range of the other three systems. Under Macro tools, where it also uses the full budget, it reaches 78.9 and 84.2, again comparable to the others. Its low scores therefore come from early stopping: when it keeps playing, it progresses as far as the other systems.

\subsection{Diagnosing Agent Limitations}
\label{sec:ablations}
\begin{figure*}[t]
\centering\begin{minipage}{\designwidth}
\centering
\includegraphics[scale=\designscale]{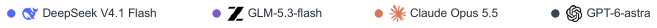}\par\vspace{1pt}
\begin{subfigure}[b]{0.230\linewidth}
\centering
\includegraphics[width=\linewidth]{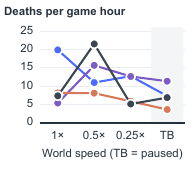}
\caption{World Speed}
\end{subfigure}\hfill
\begin{subfigure}[b]{0.228\linewidth}
\centering
\includegraphics[width=\linewidth]{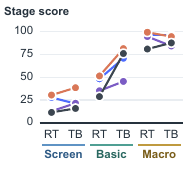}
\caption{Real-Time vs.\ Turn-Based}
\end{subfigure}\hfill
\begin{subfigure}[b]{0.230\linewidth}
\centering
\includegraphics[width=\linewidth]{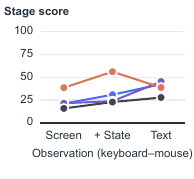}
\caption{Observation}
\end{subfigure}\hfill
\begin{subfigure}[b]{0.292\linewidth}
\centering
\includegraphics[width=\linewidth]{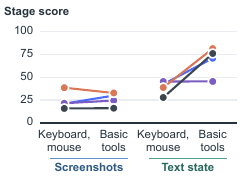}
\caption{Action}
\end{subfigure}
\caption{Ablations on the survival stages. (a) Deaths per hour of game time in Screen runs by world speed. (b) Stage scores under the real-time and turn-based clocks for the Screen interface, Basic tools and Macro tools. (c) Stage scores by observation, with keyboard--mouse control under the turn-based clock. (d) Stage scores by action under the turn-based clock, with screenshots or text observation.}
\label{fig:realtime}
\end{minipage}
\end{figure*}
We examine three possible bottlenecks behind these results: the world clock, the observation and the action, as summarized in Figure~\ref{fig:realtime}. For the \emph{world clock}, the world keeps moving while a model reasons, so we ask whether slowing or pausing the world improves performance. For the \emph{observation}, agents may fail because they cannot build an accurate picture of the world from screenshots, so we ask how performance changes when the state is given as text. For the \emph{action}, models may be poor at producing keyboard--mouse control, so we replace it with Basic-tool actions.

\subsubsection{World Clock}
\label{sec:clock-results}
To test whether agents fail because the world moves faster than they decide, we slow the world to half and quarter speed and pause it between calls. Besides scores, we report deaths per hour of game time, which measures how often an agent dies for the same exposure to the world.

More time to react reduces deaths but does not let agents solve tasks under Screen-TB. Pausing the world cuts mean deaths per run from 2.9--27.0 to 0.6--1.1, largely because less game time passes within the same budget: a paused run covers 3--19 game minutes, against 23--107 under Screen-RT. Per hour of game time, the death rate falls from 19.9 to 7.3 for DeepSeek V4.1 Flash, from 8.2 to 3.6 for Claude Opus 5.5 and from 7.4 to 6.9 for GPT-6-astra, as shown in Figure~\ref{fig:realtime}a, and GLM-5.3-flash is the exception. Pooled over all stage runs, it falls from 10.5 to 5.7. No system solves a stage at any world speed. \textbf{Overall, pausing the world lowers the number of deaths per hour of game time, but it does not let Screen agents solve a stage.}

Pausing between calls helps far more with Basic tools, which supply state and reliable actions, as shown in Figure~\ref{fig:realtime}b. Moving from Basic-RT to Basic-TB raises the stage mean of every system, and each solves two to five of the eight stages, compared with at most two under Basic-RT.

\subsubsection{Observation}
\label{sec:interface-results}
\label{sec:observation-results}
The Screen interface and Basic tools differ in both what the agent observes and how it acts. To isolate the observation, we keep keyboard--mouse control under the turn-based clock and change only the observation. Screen + State adds the player's position, orientation, health, hunger, inventory and targeted block as text beside each screenshot. A second variant replaces the screenshots with text observation.

\textbf{Agents gain when the state is given explicitly.} Adding the text readout to the screenshots raises the stage mean of every system, by 2.3 to 17.4 points, as shown in Figure~\ref{fig:realtime}c. Replacing the screenshots with text entirely raises three systems further, to 27.7--45.1. The strongest system is the exception: Claude Opus 5.5 scores 38.6 with text alone, about the same as with screenshots alone (38.4) and below its 55.8 with screenshots plus the readout. This suggests that Claude Opus 5.5 draws information from screenshots that the text does not carry, whereas the other three systems do better once the screenshots are removed. Visual perception is therefore still a bottleneck for most of the tested models.

\subsubsection{Action}
\label{sec:action-results}
To isolate the action, we replace keyboard--mouse control with the actions of the Basic tools under the turn-based clock, once for each observation, as shown in Figure~\ref{fig:realtime}d.

\textbf{With screenshots, the replacement changes stage means little.} The agent names the targets of its actions by pixel coordinates, so it still has to locate them in the screenshot, although it no longer steers the camera or times key presses. Three systems gain 0.2 to 8.8 points and Claude Opus 5.5 loses 5.8.

\textbf{With text observation, the same replacement brings large gains.} It adds 28.3, 42.4 and 48.0 points for DeepSeek V4.1 Flash, Claude Opus 5.5 and GPT-6-astra, and 0.1 for GLM-5.3-flash. The benefit of an action space is thus limited by how the agent perceives the world. \textbf{Once perception is no longer the bottleneck, tool calls are a far better way to act than keyboard--mouse control.}

\subsection{Progress and Task Mechanics}
\label{sec:behavior}
\begin{figure}[t]
\centering
\begin{subfigure}[b]{0.42\linewidth}
\centering
\includegraphics[width=\linewidth]{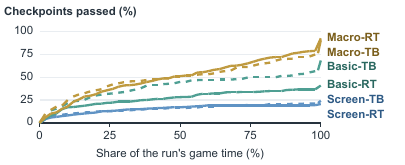}
\caption{Progress within a Stage Run}
\end{subfigure}\hspace{0.06\linewidth}
\begin{subfigure}[b]{0.42\linewidth}
\centering
\includegraphics[width=\linewidth]{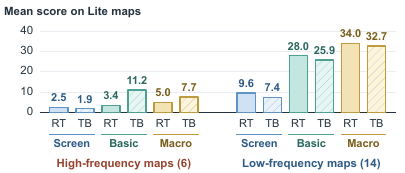}
\caption{Lite-Map Scores by Decision Frequency}
\end{subfigure}
\caption{(a) Share of a stage's checkpoints passed against each run's own elapsed game time, averaged over the stage runs of each setting. (b) Mean scores on the high- and low-frequency Lite maps, averaged over the four systems.}
\label{fig:behavior}
\end{figure}

We next compare how progress accumulates within a run under each setting. Screen agents stop making progress early. As shown in Figure~\ref{fig:behavior}a, Screen-RT agents earn 13.1 of their final 20.6 stage points within the first quarter of the run and add only 3.5 points in the second half, and 13 of the 32 runs pass no checkpoint at all. The stall is specific to the Screen interface, since Basic-RT runs gain 12.6 points in the second half and Macro runs about 40. Basic tools keep a map and notes, whereas a Screen agent must remember what it has seen on its own. \textbf{This pattern is consistent with long-horizon visual memory being a major bottleneck: agents that must remember the world through screenshots gain little after the opening.}

Under the real-time clock, tools help where a task tolerates slow decisions and not where it demands fast ones, as shown in Figure~\ref{fig:behavior}b. On the 14 low-frequency Lite maps, which center on exploration, puzzle solving and navigation, tools nearly triple the mean score from 9.6 under Screen-RT to 28.0 under Basic-RT, and pausing the clock adds nothing further (25.9 under Basic-TB). On the six high-frequency Lite maps, which involve parkour jumps, timed combat or dropper falls, every real-time setting scores at most 5 points, including Basic-RT and Macro-RT. Pausing the world raises these scores to 11.2 with Basic tools and 7.7 with Macro tools, whereas Screen-TB remains at 1.9. \textbf{On high-frequency tasks under the real-time clock, neither Basic nor Macro tools bring a clear gain, and pausing helps only modestly.}

Low scores on community maps arise for different reasons. On a parkour map, every Screen-RT run scores zero, and in the inspected runs the agents miss or die on the opening route, a failure of precise movement. On a combat map, agents die before fighting or never find the trigger that starts a wave, a failure to work out the map's rules. Community maps therefore test whether an agent can recognize and follow the rules of an unfamiliar task.

%% file: sections/table_main.tex
\colorlet{up1}{ForestGreen!50}\colorlet{up2}{ForestGreen!72}\colorlet{up3}{ForestGreen}\colorlet{up4}{ForestGreen!60!black}
\colorlet{down1}{BrickRed!50}\colorlet{down2}{BrickRed!72}\colorlet{down3}{BrickRed}\colorlet{down4}{BrickRed!60!black}
\newcommand{\dup}[2]{\,{\scriptsize\textcolor{up#1}{$\uparrow$#2}}}
\newcommand{\ddown}[2]{\,{\scriptsize\textcolor{down#1}{$\downarrow$#2}}}
\newcommand{\same}{\,{\scriptsize\textcolor{gray}{0.0}}}
\begin{table*}[t]
\centering\fontsize{7.8}{9.8}\selectfont
\caption{Main results on \benchmark{}-Lite. GPT-6-astra runs in Codex and the other models in Claude Code. Score is the mean task score from 0 to 100, and the \benchmark{}-Lite score is the mean over all 29 tasks; Solved counts full-credit runs; Deaths and Turns are per-run means. Icons and row tints mark the interface of each setting, either screen in blue, Basic tools in teal or Macro tools in sand, and icons mark its clock, either real time with a lightning bolt or turn-based with a pause sign. Bold marks the best value per setting. Arrows show changes from the same system's Screen-RT result: green for increases, red for decreases, and darker shades for larger changes within a column.}
\label{tab:main-results}
\vspace{2pt}
\setlength{\tabcolsep}{2.5pt}\renewcommand{\arraystretch}{1.0}
\begin{tabular}{l l cccc cccc c c}
\toprule
\lchead & & \multicolumn{4}{c}{Survival stages (8)} & \multicolumn{4}{c}{Lite maps (20)} & Mainline (1) & \benchmark{}-Lite (29) \\
\cmidrule(lr){3-6}\cmidrule(lr){7-10}\cmidrule(lr){11-11}\cmidrule(lr){12-12}
\lchead Setting & Model & Score & Solved & Deaths & Turns & Score & Solved & Deaths & Turns & Score & Score \\
\midrule
\rowcolor{LCBlueFill} & \ico{deepseek}\,DeepSeek V4.1 Flash & 27.7 & 0 & 27.0 & 162 & 5.9 & 0 & 1.5 & 263 & 15.8 & 12.2 \\
\rowcolor{LCBlueFill} & \ico{glm}\,GLM-5.3-flash & 12.8 & 0 & 9.6 & 145 & 7.4 & 0 & 3.4 & 228 & 17.1 & 9.2 \\
\rowcolor{LCBlueFill} & \ico{claude}\,Claude Opus 5.5 & \textbf{30.3} & 0 & 8.6 & 188 & \textbf{10.8} & 0 & 4.1 & 207 & \textbf{32.9} & \textbf{17.0} \\
\rowcolor{LCBlueFill}\multirow{-4}{*}{\shortstack[l]{\ico{screen}\,\ico{rt}\\[1pt]Screen-RT}} & \ico{openai}\,GPT-6-astra & 11.5 & 0 & \textbf{2.9} & 71 & 6.0 & 0 & \textbf{1.2} & 49 & 1.3 & 7.3 \\
\midrule
\rowcolor{LCBlueFill} & \ico{deepseek}\,DeepSeek V4.1 Flash & 21.0\ddown{1}{6.7} & 0 & 0.9\ddown{4}{26.1} & 207 & \textbf{7.4}\dup{1}{1.5} & 0 & 2.2\dup{1}{0.7} & 290 & 19.7\dup{1}{3.9} & 11.5\ddown{1}{0.7} \\
\rowcolor{LCBlueFill} & \ico{glm}\,GLM-5.3-flash & 21.4\dup{1}{8.6} & 0 & \textbf{0.6}\ddown{2}{9.0} & 178 & 4.9\ddown{1}{2.5} & 0 & 1.4\ddown{2}{2.0} & 255 & 17.1\same & 9.9\dup{1}{0.7} \\
\rowcolor{LCBlueFill} & \ico{claude}\,Claude Opus 5.5 & \textbf{38.4}\dup{1}{8.1} & 0 & 1.1\ddown{2}{7.5} & 358 & 5.4\ddown{1}{5.4} & 0 & 3.4\ddown{1}{0.7} & 290 & \textbf{40.8}\dup{1}{7.9} & \textbf{15.7}\ddown{1}{1.3} \\
\rowcolor{LCBlueFill}\multirow{-4}{*}{\shortstack[l]{\ico{screen}\,\ico{tb}\\[1pt]Screen-TB}} & \ico{openai}\,GPT-6-astra & 15.8\dup{1}{4.3} & 0 & 0.8\ddown{1}{2.1} & 75 & 5.3\ddown{1}{0.7} & 0 & \textbf{0.8}\ddown{1}{0.4} & 53 & 15.8\dup{1}{14.5} & 8.5\dup{1}{1.2} \\
\midrule
\rowcolor{LCTealFill} & \ico{deepseek}\,DeepSeek V4.1 Flash & 48.0\dup{1}{20.3} & \textbf{2} & 25.3\ddown{1}{1.7} & 758 & 19.0\dup{2}{13.1} & 1 & 3.6\dup{3}{2.1} & 651 & 43.4\dup{2}{27.6} & 27.8\dup{2}{15.6} \\
\rowcolor{LCTealFill} & \ico{glm}\,GLM-5.3-flash & 35.0\dup{2}{22.2} & 0 & 19.3\dup{2}{9.7} & 720 & 14.7\dup{1}{7.3} & 0 & 6.9\dup{4}{3.5} & 458 & 39.5\dup{2}{22.4} & 21.1\dup{2}{11.9} \\
\rowcolor{LCTealFill} & \ico{claude}\,Claude Opus 5.5 & \textbf{51.2}\dup{2}{20.9} & \textbf{2} & 6.3\ddown{1}{2.3} & 387 & \textbf{28.4}\dup{3}{17.6} & 1 & 3.8\ddown{1}{0.3} & 383 & \textbf{48.7}\dup{1}{15.8} & \textbf{35.4}\dup{2}{18.4} \\
\rowcolor{LCTealFill}\multirow{-4}{*}{\shortstack[l]{\ico{basic}\,\ico{rt}\\[1pt]Basic-RT}} & \ico{openai}\,GPT-6-astra & 28.6\dup{1}{17.1} & 0 & \textbf{4.9}\dup{1}{2.0} & 94 & 20.5\dup{2}{14.5} & \textbf{2} & \textbf{2.3}\dup{2}{1.1} & 156 & 7.9\dup{1}{6.6} & 22.3\dup{2}{15.0} \\
\midrule
\rowcolor{LCTealFill} & \ico{deepseek}\,DeepSeek V4.1 Flash & 70.5\dup{3}{42.8} & 3 & 1.9\ddown{4}{25.1} & 583 & 20.3\dup{2}{14.4} & 1 & 3.3\dup{2}{1.8} & 794 & 44.7\dup{2}{28.9} & 35.0\dup{3}{22.8} \\
\rowcolor{LCTealFill} & \ico{glm}\,GLM-5.3-flash & 45.2\dup{2}{32.4} & 2 & 2.5\ddown{2}{7.1} & 638 & 14.5\dup{1}{7.1} & 0 & 4.3\dup{1}{0.9} & 538 & 39.5\dup{2}{22.4} & 23.8\dup{2}{14.6} \\
\rowcolor{LCTealFill} & \ico{claude}\,Claude Opus 5.5 & \textbf{81.0}\dup{3}{50.7} & \textbf{5} & 2.6\ddown{1}{6.0} & 349 & \textbf{31.1}\dup{3}{20.3} & \textbf{3} & 2.0\ddown{3}{2.1} & 343 & \textbf{59.2}\dup{2}{26.3} & \textbf{45.8}\dup{3}{28.8} \\
\rowcolor{LCTealFill}\multirow{-4}{*}{\shortstack[l]{\ico{basic}\,\ico{tb}\\[1pt]Basic-TB}} & \ico{openai}\,GPT-6-astra & 75.7\dup{4}{64.2} & 4 & \textbf{0.9}\ddown{1}{2.0} & 74 & 20.0\dup{2}{14.0} & \textbf{3} & \textbf{1.2}\same & 54 & 43.4\dup{3}{42.1} & 36.1\dup{3}{28.8} \\
\midrule
\rowcolor{LCSandFill} & \ico{deepseek}\,DeepSeek V4.1 Flash & 95.5\dup{4}{67.8} & \textbf{7} & 0.4\ddown{4}{26.6} & 115 & 22.3\dup{3}{16.4} & 2 & 5.6\dup{4}{4.1} & 514 & 77.6\dup{3}{61.8} & 44.4\dup{4}{32.2} \\
\rowcolor{LCSandFill} & \ico{glm}\,GLM-5.3-flash & 94.2\dup{4}{81.4} & 6 & 0.6\ddown{2}{9.0} & 179 & 18.5\dup{2}{11.1} & 2 & 1.5\ddown{2}{1.9} & 468 & 65.8\dup{3}{48.7} & 41.0\dup{4}{31.8} \\
\rowcolor{LCSandFill} & \ico{claude}\,Claude Opus 5.5 & \textbf{98.9}\dup{4}{68.6} & \textbf{7} & \textbf{0.3}\ddown{2}{8.3} & 48 & \textbf{39.6}\dup{4}{28.8} & \textbf{3} & 1.7\ddown{3}{2.4} & 286 & \textbf{89.5}\dup{3}{56.6} & \textbf{57.7}\dup{4}{40.7} \\
\rowcolor{LCSandFill}\multirow{-4}{*}{\shortstack[l]{\ico{macro}\,\ico{rt}\\[1pt]Macro-RT}} & \ico{openai}\,GPT-6-astra & 80.6\dup{4}{69.1} & 5 & \textbf{0.3}\ddown{1}{2.6} & 34 & 20.9\dup{3}{14.9} & 2 & \textbf{0.5}\ddown{1}{0.7} & 39 & 78.9\dup{4}{77.6} & 39.3\dup{4}{32.0} \\
\midrule
\rowcolor{LCSandFill} & \ico{deepseek}\,DeepSeek V4.1 Flash & \textbf{95.5}\dup{4}{67.8} & \textbf{7} & 0.9\ddown{4}{26.1} & 115 & 18.8\dup{2}{12.9} & 1 & 2.4\dup{1}{0.9} & 542 & 65.8\dup{3}{50.0} & 41.6\dup{3}{29.4} \\
\rowcolor{LCSandFill} & \ico{glm}\,GLM-5.3-flash & 83.8\dup{4}{71.0} & 4 & 3.0\ddown{1}{6.6} & 138 & 16.4\dup{2}{9.0} & 1 & 1.2\ddown{3}{2.2} & 429 & 63.2\dup{3}{46.1} & 36.6\dup{3}{27.4} \\
\rowcolor{LCSandFill} & \ico{claude}\,Claude Opus 5.5 & 94.1\dup{4}{63.8} & 6 & 0.4\ddown{2}{8.2} & 27 & \textbf{40.2}\dup{4}{29.4} & \textbf{3} & 1.0\ddown{4}{3.1} & 242 & \textbf{96.1}\dup{4}{63.2} & \textbf{57.0}\dup{4}{40.0} \\
\rowcolor{LCSandFill}\multirow{-4}{*}{\shortstack[l]{\ico{macro}\,\ico{tb}\\[1pt]Macro-TB}} & \ico{openai}\,GPT-6-astra & 87.2\dup{4}{75.7} & 5 & \textbf{0.3}\ddown{1}{2.6} & 35 & 25.4\dup{3}{19.4} & \textbf{3} & \textbf{0.9}\ddown{1}{0.3} & 65 & 84.2\dup{4}{82.9} & 44.4\dup{4}{37.1} \\
\bottomrule
\end{tabular}

\end{table*}

%% file: sections/results_figure.tex
\begin{figure*}[t]
\centering\begin{minipage}{\designwidth}
\centering
\includegraphics[scale=\designscale]{figures/legend_systems.pdf}\par\vspace{1pt}
\begin{subfigure}[b]{0.36\linewidth}
\centering
\includegraphics[width=\linewidth]{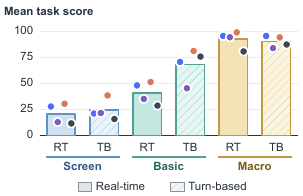}
\caption{Survival Stages}
\end{subfigure}\hfill
\begin{subfigure}[b]{0.33\linewidth}
\centering
\includegraphics[width=\linewidth]{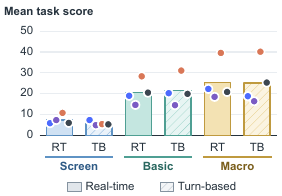}
\caption{Lite Maps}
\end{subfigure}\hfill
\begin{subfigure}[b]{0.29\linewidth}
\centering
\includegraphics[width=\linewidth]{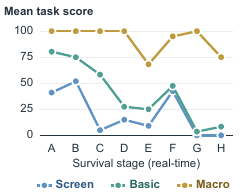}
\caption{Survival Stages by Stage}
\end{subfigure}
\caption{Scores by setting and stage, one run per task. (a) Survival-stage means, with bars averaging the four systems and markers showing each system. (b) Means on the 20 Lite maps. (c) Stage scores under the real-time clock by stage, averaged over the four systems.}
\label{fig:results}
\end{minipage}
\end{figure*}

%% file: sections/conclusion.tex
\section{Conclusion}
\label{sec:conclusion}
Real-time open worlds require agents to retain useful information and keep acting toward a goal while their surroundings change. To evaluate these capabilities, we build \benchmark{} from initialized Minecraft survival stages, a continuous mainline and diverse community maps, with native screenshot control and code-based progress checks. None of four frontier agent systems solves any \benchmark{}-Lite task under Screen-RT. Pausing the world tends to lower death rates per hour of game time but yields no solved stage with the Screen interface, while survival-oriented Macro tools nearly solve the stages but add far less on community maps. These findings call for agents that perceive and act reliably from pixels and sustain progress over long horizons while the world keeps running.

%% file: sections/appendix_results.tex
\section{Additional Results and Analysis}
\label{app:results}
\subsection{Overall Results}
\label{app:overall}
\label{app:scores}
\paragraph{\benchmark{}-Full.}
Table~\ref{tab:full-score} extends the Screen-RT evaluation from \benchmark{}-Lite to all 79 tasks. Across \benchmark{}-Full, the four systems score between 4.0 and 12.4, below their \benchmark{}-Lite scores of 7.3 to 17.0, because the community maps, on which every system scores well below its stage mean, make up a larger share of the tasks. The relative ordering of the four systems matches \benchmark{}-Lite. Across all 70 community maps, mean scores range from 3.2 to 10.1, with 10 to 23 maps per system earning any credit, and only Claude Opus 5.5 solves any map (3 of 70). GPT-6-astra terminates 53 of its 70 map runs before the time limit, consistent with the early-stopping pattern analyzed in Section~\ref{sec:overall}.

\begin{table}[!htbp]
\centering\small
\caption{\benchmark{}-Full results under Screen-RT. The first four columns are mean scores per task group, and the last four cover all 70 community maps.}
\label{tab:full-score}
\vspace{2pt}
\setlength{\tabcolsep}{3.2pt}
\begin{tabularx}{\linewidth}{l RRRR RR RRRR}
\toprule
\lchead & \multicolumn{4}{c}{Task group} & \multicolumn{2}{c}{Benchmark score} & \multicolumn{4}{c}{All 70 community maps} \\
\cmidrule(lr){2-5}\cmidrule(lr){6-7}\cmidrule(l){8-11}
\lchead System & Stages & \shortstack{Lite\\maps} & \shortstack{Other\\maps} & \shortstack{Main-\\line} & Lite & Full & Score & Solved & \shortstack{With\\credit} & \shortstack{Agent\\stops} \\
\midrule
DeepSeek V4.1 Flash & 27.7 & 5.9 & 5.8 & 15.8 & 12.2 & 8.1 & 5.8 & 0 & 17 & 13 \\
GLM-5.3-flash & 12.8 & 7.4 & 5.1 & 17.1 & 9.2 & 6.6 & 5.8 & 0 & 16 & 2 \\
Claude Opus 5.5 & \textbf{30.3} & \textbf{10.8} & \textbf{9.8} & \textbf{32.9} & \textbf{17.0} & \textbf{12.4} & \textbf{10.1} & \textbf{3} & \textbf{23} & 26 \\
GPT-6-astra & 11.5 & 6.0 & 2.1 & 1.3 & 7.3 & 4.0 & 3.2 & 0 & 10 & 53 \\
\bottomrule
\end{tabularx}
\end{table}

\paragraph{Scores by stage.}
Table~\ref{tab:stage-visual} reports per-stage scores for the Screen interface under both clocks, expanding the stage means in Table~\ref{tab:main-results}. Credit concentrates in stages A, B and F under both clocks, and pausing the world yields credit on stage G for only one system. On stage F, the three systems reaching 50.0 under Screen-RT satisfy only the five local checks for crafting blaze powder and crafting and throwing Eyes of Ender, while no run satisfies a stronghold checkpoint.

\begin{table}[!htbp]
\centering\small
\caption{Survival-stage scores of each system under Screen-RT and Screen-TB.}
\label{tab:stage-visual}
\vspace{2pt}
\setlength{\tabcolsep}{4.5pt}
\begin{tabularx}{\linewidth}{l RR RR RR RR}
\toprule
\lchead & \multicolumn{2}{c}{DeepSeek V4.1 Flash} & \multicolumn{2}{c}{GLM-5.3-flash} & \multicolumn{2}{c}{Claude Opus 5.5} & \multicolumn{2}{c}{GPT-6-astra} \\
\cmidrule(lr){2-3}\cmidrule(lr){4-5}\cmidrule(lr){6-7}\cmidrule(l){8-9}
\lchead Stage & RT & TB & RT & TB & RT & TB & RT & TB \\
\midrule
A: Wood and shelter & 50.0 & 50.0 & 21.4 & 42.9 & 64.3 & 64.3 & 28.6 & 28.6 \\
B: Stone tools and food & 76.9 & 23.1 & 30.8 & 53.8 & 76.9 & 76.9 & 23.1 & 30.8 \\
C: Iron tools & 6.7 & 6.7 & 0.0 & 6.7 & 13.3 & 53.3 & 0.0 & 6.7 \\
D: Diamond mining & 20.0 & 20.0 & 0.0 & 0.0 & 20.0 & 20.0 & 20.0 & 10.0 \\
E: Nether resources & 18.2 & 18.2 & 0.0 & 18.2 & 18.2 & 18.2 & 0.0 & 0.0 \\
F: Stronghold search & 50.0 & 50.0 & 50.0 & 50.0 & 50.0 & 60.0 & 20.0 & 50.0 \\
G: End-portal access & 0.0 & 0.0 & 0.0 & 0.0 & 0.0 & 14.3 & 0.0 & 0.0 \\
H: Dragon fight & 0.0 & 0.0 & 0.0 & 0.0 & 0.0 & 0.0 & 0.0 & 0.0 \\
\midrule
Mean & 27.7 & 21.0 & 12.8 & 21.4 & 30.3 & 38.4 & 11.5 & 15.8 \\
\bottomrule
\end{tabularx}
\end{table}

\paragraph{Run endings and turns.}
\label{app:endings}
\label{app:longhorizon}
We classify termination outcomes for each stage and Lite-map run from its execution record using the first matching condition in precedence order: verifier termination at the death limit, map-specific terminal failure (such as the first death on Apocalyptic), satisfaction of all checkpoints, wall-clock budget expiration, or early session termination by the agent. Runs whose agent process exits with an error, or whose ending is not recorded, are counted under Other. Turns match the model-turn counts in Table~\ref{tab:main-results} and include all assistant responses across context compactions. As summarized in Table~\ref{tab:turns-stops}, termination behavior varies markedly across systems: GPT-6-astra terminates most of its runs before the budget expires, GLM-5.3-flash predominantly reaches the wall-clock limit, and death-limit terminations occur almost exclusively in real-time stage runs. Model-turn counts also vary widely across interfaces and systems: Basic stage runs of DeepSeek V4.1 Flash and GLM-5.3-flash average 583 to 758 turns, whereas Macro stage runs of Claude Opus 5.5 and GPT-6-astra average 27 to 48 turns because individual Macro calls encapsulate multi-step routines.

\begin{table}[!htbp]
\centering\footnotesize
\caption{Run endings and model turns by system and setting. Ending columns count runs, and Compactions is the mean number of context compactions per run.}
\label{tab:turns-stops}
\vspace{2pt}
\setlength{\tabcolsep}{2.4pt}
\resizebox{\linewidth}{!}{%
\begin{tabular}{l l rrrrr rr rrrrrr rr}
\toprule
\lchead & & \multicolumn{7}{c}{Survival stages (8)} & \multicolumn{8}{c}{Lite maps (20)} \\
\cmidrule(lr){3-9}\cmidrule(l){10-17}
\lchead System & Setting & Solved & \shortstack{Time\\limit} & \shortstack{Agent\\stop} & \shortstack{Death\\limit} & Other & Turns & \shortstack{Compac-\\tions} & Solved & \shortstack{Time\\limit} & \shortstack{Agent\\stop} & \shortstack{Death\\limit} & \shortstack{Task\\failure} & Other & Turns & \shortstack{Compac-\\tions} \\
\midrule
\rowcolor{LCBlueFill}\cellcolor{white} & Screen-RT & 0 & 2 & 3 & 3 & 0 & 162 & 0.8 & 0 & 12 & 3 & 0 & 5 & 0 & 263 & 1.8 \\
\rowcolor{LCBlueFill}\cellcolor{white} & Screen-TB & 0 & 7 & 1 & 0 & 0 & 207 & 1.8 & 0 & 13 & 2 & 0 & 5 & 0 & 290 & 1.9 \\
\rowcolor{LCTealFill}\cellcolor{white} & Basic-RT & 2 & 4 & 0 & 2 & 0 & 758 & 4.5 & 1 & 12 & 2 & 0 & 5 & 0 & 651 & 6.2 \\
\rowcolor{LCTealFill}\cellcolor{white} & Basic-TB & 3 & 3 & 2 & 0 & 0 & 583 & 4.4 & 1 & 13 & 2 & 0 & 4 & 0 & 794 & 8.3 \\
\rowcolor{LCSandFill}\cellcolor{white} & Macro-RT & 7 & 1 & 0 & 0 & 0 & 115 & 0.9 & 2 & 10 & 5 & 0 & 3 & 0 & 514 & 9.4 \\
\rowcolor{LCSandFill}\cellcolor{white}\multirow{-6}{*}{\shortstack[l]{DeepSeek\\V4.1 Flash}} & Macro-TB & 7 & 1 & 0 & 0 & 0 & 115 & 1.2 & 1 & 8 & 6 & 0 & 5 & 0 & 542 & 8.7 \\
\midrule
\rowcolor{LCBlueFill}\cellcolor{white} & Screen-RT & 0 & 7 & 0 & 1 & 0 & 145 & 0.4 & 0 & 15 & 0 & 0 & 5 & 0 & 228 & 0.7 \\
\rowcolor{LCBlueFill}\cellcolor{white} & Screen-TB & 0 & 8 & 0 & 0 & 0 & 178 & 0.4 & 0 & 16 & 1 & 0 & 3 & 0 & 255 & 1.0 \\
\rowcolor{LCTealFill}\cellcolor{white} & Basic-RT & 0 & 6 & 0 & 2 & 0 & 720 & 2.6 & 0 & 13 & 1 & 1 & 5 & 0 & 458 & 2.6 \\
\rowcolor{LCTealFill}\cellcolor{white} & Basic-TB & 2 & 6 & 0 & 0 & 0 & 638 & 2.2 & 0 & 14 & 1 & 0 & 5 & 0 & 538 & 2.9 \\
\rowcolor{LCSandFill}\cellcolor{white} & Macro-RT & 6 & 1 & 1 & 0 & 0 & 179 & 0.6 & 2 & 14 & 1 & 0 & 3 & 0 & 468 & 3.2 \\
\rowcolor{LCSandFill}\cellcolor{white}\multirow{-6}{*}{GLM-5.3-flash} & Macro-TB & 4 & 2 & 2 & 0 & 0 & 138 & 0.6 & 1 & 14 & 1 & 0 & 4 & 0 & 429 & 3.5 \\
\midrule
\rowcolor{LCBlueFill}\cellcolor{white} & Screen-RT & 0 & 3 & 4 & 1 & 0 & 188 & 1.8 & 0 & 9 & 6 & 0 & 5 & 0 & 207 & 0.9 \\
\rowcolor{LCBlueFill}\cellcolor{white} & Screen-TB & 0 & 7 & 1 & 0 & 0 & 358 & 3.5 & 0 & 11 & 5 & 0 & 4 & 0 & 290 & 1.1 \\
\rowcolor{LCTealFill}\cellcolor{white} & Basic-RT & 2 & 3 & 3 & 0 & 0 & 387 & 3.8 & 1 & 9 & 5 & 0 & 5 & 0 & 383 & 3.6 \\
\rowcolor{LCTealFill}\cellcolor{white} & Basic-TB & 5 & 2 & 1 & 0 & 0 & 349 & 3.2 & 3 & 8 & 5 & 0 & 3 & 1 & 343 & 3.2 \\
\rowcolor{LCSandFill}\cellcolor{white} & Macro-RT & 7 & 1 & 0 & 0 & 0 & 48 & 0.4 & 3 & 8 & 3 & 0 & 4 & 2 & 286 & 4.4 \\
\rowcolor{LCSandFill}\cellcolor{white}\multirow{-6}{*}{\shortstack[l]{Claude\\Opus 5.5}} & Macro-TB & 6 & 1 & 1 & 0 & 0 & 27 & 0.0 & 3 & 5 & 6 & 0 & 3 & 3 & 242 & 3.4 \\
\midrule
\rowcolor{LCBlueFill}\cellcolor{white} & Screen-RT & 0 & 0 & 8 & 0 & 0 & 71 & 0.0 & 0 & 2 & 13 & 0 & 5 & 0 & 49 & 0.0 \\
\rowcolor{LCBlueFill}\cellcolor{white} & Screen-TB & 0 & 0 & 8 & 0 & 0 & 75 & 0.0 & 0 & 3 & 13 & 0 & 4 & 0 & 53 & 0.0 \\
\rowcolor{LCTealFill}\cellcolor{white} & Basic-RT & 0 & 1 & 7 & 0 & 0 & 94 & 0.0 & 2 & 3 & 11 & 0 & 4 & 0 & 156 & 0.2 \\
\rowcolor{LCTealFill}\cellcolor{white} & Basic-TB & 4 & 0 & 4 & 0 & 0 & 74 & 0.0 & 3 & 1 & 14 & 0 & 2 & 0 & 54 & 0.0 \\
\rowcolor{LCSandFill}\cellcolor{white} & Macro-RT & 5 & 0 & 3 & 0 & 0 & 34 & 0.0 & 2 & 0 & 12 & 0 & 4 & 2 & 39 & 0.0 \\
\rowcolor{LCSandFill}\cellcolor{white}\multirow{-6}{*}{GPT-6-astra} & Macro-TB & 5 & 0 & 2 & 0 & 1 & 35 & 0.0 & 3 & 1 & 13 & 0 & 3 & 0 & 65 & 0.1 \\
\bottomrule
\end{tabular}}
\end{table}

\paragraph{Early stopping by GPT-6-astra.}
GPT-6-astra frequently terminates its sessions while most of the wall-clock budget remains. Across its 168 stage and Lite-map runs, it ends 108 sessions early at a median elapsed time of 15 minutes (12\% of the budget), with 63 of those runs scoring zero. All 16 of its Screen stage runs terminate early, after 5 to 56 of the 120 available minutes. On the continuous mainline, it terminates its Screen-RT and Basic-RT runs after 161 and 31 of the 600 available minutes, while utilizing the full ten-hour budget in the other four settings. The representative final messages in Table~\ref{tab:astra-stops} summarize partial progress and report the objective as incomplete, in several cases phrasing the run as already concluded (``before the run ended'').

\begin{table}[!htbp]
\centering\footnotesize
\caption{Final messages of GPT-6-astra in five self-terminated runs, quoted from the agent logs and shortened where marked. Time is the wall-clock time used out of the budget.}
\label{tab:astra-stops}
\vspace{2pt}
\setlength{\tabcolsep}{4pt}
\begin{tabularx}{\linewidth}{l l r r >{\raggedright\arraybackslash}X}
\toprule
\lchead Setting & Task & Time (min) & Score & Final message \\
\midrule
Screen-RT & Stage C & 42 / 120 & 0.0 & ``I mined and collected raw iron underground, but could not complete smelting and crafting before the run ended.'' \\
Screen-RT & Stage G & 31 / 120 & 0.0 & ``I wasn't able to complete the portal activation run.'' \\
Screen-TB & Stage E & 17 / 120 & 0.0 & ``Unable to complete the Nether objectives in this run.'' \\
Screen-TB & Stage H & 16 / 120 & 0.0 & ``I wasn't able to complete the End fight objectives.'' \\
Macro-RT & Dropper map & 22 / 120 & 0.0 & ``No milestones completed. Two falls caused deaths [\ldots] I stopped rather than risk the remaining three lives.'' \\
\bottomrule
\end{tabularx}
\end{table}

\FloatBarrier
\subsection{World Clock}
\label{app:clock-all}
\label{app:clock}
Table~\ref{tab:realtime} reports the complete per-speed metrics underlying Figure~\ref{fig:realtime}a (Section~\ref{sec:clock-results}), including stage score, deaths per run, deaths per game hour, elapsed game time, and inter-call game time across full speed (1$\times$), half speed (0.5$\times$), quarter speed (0.25$\times$), and paused (Screen-TB) execution.

\begin{table}[!htbp]
\centering\small
\caption{Screen stage runs by world speed. The 1$\times$ column is Screen-RT and the Paused column is Screen-TB. Game minutes are the mean game time per run, and game seconds between calls the mean game time that passes between two calls.}
\label{tab:realtime}
\label{tab:speed}
\vspace{2pt}
\setlength{\tabcolsep}{4pt}
\begin{tabularx}{\linewidth}{l l RRR R}
\toprule
\lchead & & \multicolumn{3}{c}{Running world, by speed} & Paused \\
\cmidrule(lr){3-5}\cmidrule(l){6-6}
\lchead System & Measure & 1$\times$ & 0.5$\times$ & 0.25$\times$ & \\
\midrule
DeepSeek V4.1 Flash & Score & 27.7 & 23.9 & 25.0 & 21.0 \\
 & Deaths per run & 27.0 & 8.4 & 5.1 & 0.9 \\
 & Deaths per game hour & 19.9 & 11.0 & 12.8 & 7.3 \\
 & Game minutes per run & 81 & 46 & 24 & 7 \\
 & Game seconds between calls & 28.7 & 12.9 & 3.2 & 0 \\
\midrule
GLM-5.3-flash & Score & 12.8 & 23.9 & 22.9 & 21.4 \\
 & Deaths per run & 9.6 & 13.3 & 6.3 & 0.6 \\
 & Deaths per game hour & 5.4 & 15.7 & 12.7 & 11.4 \\
 & Game minutes per run & 107 & 51 & 30 & 3 \\
 & Game seconds between calls & 44.0 & 13.7 & 3.8 & 0 \\
\midrule
Claude Opus 5.5 & Score & 30.3 & 34.5 & 34.0 & 38.4 \\
 & Deaths per run & 8.6 & 5.3 & 2.1 & 1.1 \\
 & Deaths per game hour & 8.2 & 8.1 & 5.9 & 3.6 \\
 & Game minutes per run & 63 & 39 & 22 & 19 \\
 & Game seconds between calls & 14.5 & 8.7 & 4.3 & 0 \\
\midrule
GPT-6-astra & Score & 11.5 & 15.5 & 16.6 & 15.8 \\
 & Deaths per run & 2.9 & 11.5 & 1.5 & 0.8 \\
 & Deaths per game hour & 7.4 & 21.5 & 5.1 & 6.9 \\
 & Game minutes per run & 23 & 32 & 17 & 7 \\
 & Game seconds between calls & 14.0 & 74.2 & 38.1 & 0 \\
\bottomrule
\end{tabularx}
\end{table}

\paragraph{Deaths and decision time.}
Deaths per game hour are computed by dividing total deaths in a cell by total elapsed game hours, $\sum_i d_i/(\sum_i g_i/72000)$, where $d_i$ and $g_i$ denote the death count and game ticks of run $i$. Normalizing by game time rather than wall-clock time avoids artificially lowering death rates when the world is slowed or paused. Real-time deaths concentrate in a small number of runs: three DeepSeek V4.1 Flash runs reach the 50-death limit and account for 150 of its 216 Screen-RT deaths. Pooled across the four systems, pausing lowers the death rate from 10.5 to 5.7 deaths per game hour (385 deaths in 36.6 game hours versus 27 in 4.7), whereas slowing the world produces inconsistent changes across systems, and no slowed run solves a stage. Under Screen-RT, model calls average 14 to 45 wall-clock seconds, leaving 14 to 44 seconds of game time between consecutive actions at 1$\times$ speed.

\paragraph{GPT-6-astra at reduced speed.}
GPT-6-astra's values in Table~\ref{tab:realtime} do not change steadily with world speed, mainly because it ends its runs itself at varying times. At full speed it ends all eight runs after 5 to 42 minutes, at half speed it keeps playing for 29 to 106 minutes before stopping, and at quarter speed three runs reach the two-hour limit. The game time that a run covers, and with it the number of deaths, therefore depends more on when the model stops than on the world speed: a half-speed run covers 32 game minutes, more than the 23 of a full-speed run. At half speed, 90 of its 92 deaths fall in stages F, G and H, where the long gaps between its calls leave the player standing among hostile mobs.

\FloatBarrier
\subsection{Observation and Action}
\label{app:obs-action-all}
\label{app:obs-action}
Table~\ref{tab:obs-action} reports stage means for the five turn-based observation--action combinations evaluated across all four systems (Sections~\ref{sec:observation-results} and~\ref{sec:action-results}), spanning from Screen-TB in the first row to Basic-TB in the last.

\begin{table}[!htbp]
\centering\small
\caption{Stage means of the observation and action variants under the turn-based clock. The first row is Screen-TB, the second Screen + State and the last Basic-TB.}
\label{tab:obs-action}
\vspace{2pt}
\setlength{\tabcolsep}{4pt}
\begin{tabularx}{\linewidth}{l l RRRR}
\toprule
\lchead Observation & Action & \shortstack[r]{DeepSeek\\V4.1 Flash} & GLM-5.3-flash & Claude Opus 5.5 & GPT-6-astra \\
\midrule
Screenshots & Keyboard and mouse & 21.0 & 21.4 & 38.4 & 15.8 \\
Screenshots + state & Keyboard and mouse & 30.7 & 23.7 & 55.8 & 22.7 \\
Text observation & Keyboard and mouse & 42.2 & 45.1 & 38.6 & 27.7 \\
Screenshots & Basic-tool actions & 29.8 & 24.4 & 32.6 & 16.0 \\
Text observation & Basic-tool actions & 70.5 & 45.2 & 81.0 & 75.7 \\
\bottomrule
\end{tabularx}
\end{table}

\paragraph{Variant specifications.}
In the Screen interface, screenshots are captured from the agent's own game client with the native heads-up display: the hotbar, health and hunger bars are visible, and the agent opens the inventory with a key and clicks its slots by pixel coordinates, as illustrated in Figure~\ref{fig:teaser}. Screen + State keeps this interface and appends a text readout of the player's position, orientation, health, hunger, inventory and targeted block to each screenshot, which includes values not visible on screen, such as coordinates and the full inventory while it is closed. The variant combining text observation with keyboard--mouse control keeps the text observation of the Basic tools but replaces all action tools with a single \texttt{act} tool that holds keys, rotates the camera and clicks inventory slots, following the conventions of the Screen interface.

The variant combining screenshots with Basic-tool actions runs in the environment of the Basic tools, where the player is a bot on the game server that has no game client of its own. Its $960\times540$ screenshot is therefore rendered by the spectator client that records the run: it follows the bot's view with the same 70-degree field of view but cannot show the bot's heads-up display. A strip with health, hunger, air and the hotbar item names is drawn along the bottom of the image to replace it. The agent names the target of an action by pixel coordinates in the most recent screenshot, and the server casts a ray to the first block or entity at that pixel. Inventory slots are not clicked in this variant: crafting, equipping and containers go through the Basic tools by item name, and the inventory, chest and furnace tools return their contents as grid images.

\paragraph{Observation.}
Adding an explicit readout of the player's state improves every system: Screen + State raises the stage mean of all four systems under both clocks, with the largest turn-based gain for Claude Opus 5.5, which under the turn-based clock also solves stage B and becomes the only system to earn credit on stage G. Replacing screenshots entirely with text observation further improves the other three systems, whereas Claude Opus 5.5 scores about the same with text alone as with screenshots alone and achieves its best result when combining screenshots with the state readout. This pattern suggests that Claude Opus 5.5 extracts spatial layout and visual cues from screenshots that a list of blocks and entities does not capture, whereas the other three systems benefit more from explicit textual state than from raw images.

\paragraph{Action.}
The benefit of Basic-tool actions depends on the observation modality. Paired with text observation, they substantially increase stage means for three systems while leaving GLM-5.3-flash nearly unchanged. Paired with screenshots, they change stage means little, and Claude Opus 5.5 scores lower than under keyboard--mouse control. One plausible explanation is interface alignment: native screenshots pair naturally with direct keyboard--mouse control, whereas pairing screenshots with Basic-tool actions introduces indirection through a spectator view without the native heads-up display, pixel-to-ray target resolution and separate grid images for containers. Only combining text observation with Basic-tool actions (Basic-TB) raises three of the four systems above 70.

\FloatBarrier
\subsection{Progress and Task Mechanics}
\label{app:progress-all}
\label{sec:progress-results}
\label{sec:mechanics-results}
\paragraph{Decision frequency.}
We group the parkour, dropper and combat categories as high-frequency because their tasks require timed jumps, descents or combat, and the remaining seven categories as low-frequency. This partition yields 6 high- and 14 low-frequency Lite maps, and 29 and 41 maps across all 70 community maps.

\begin{figure*}[t]
\centering
\includegraphics[width=\textwidth]{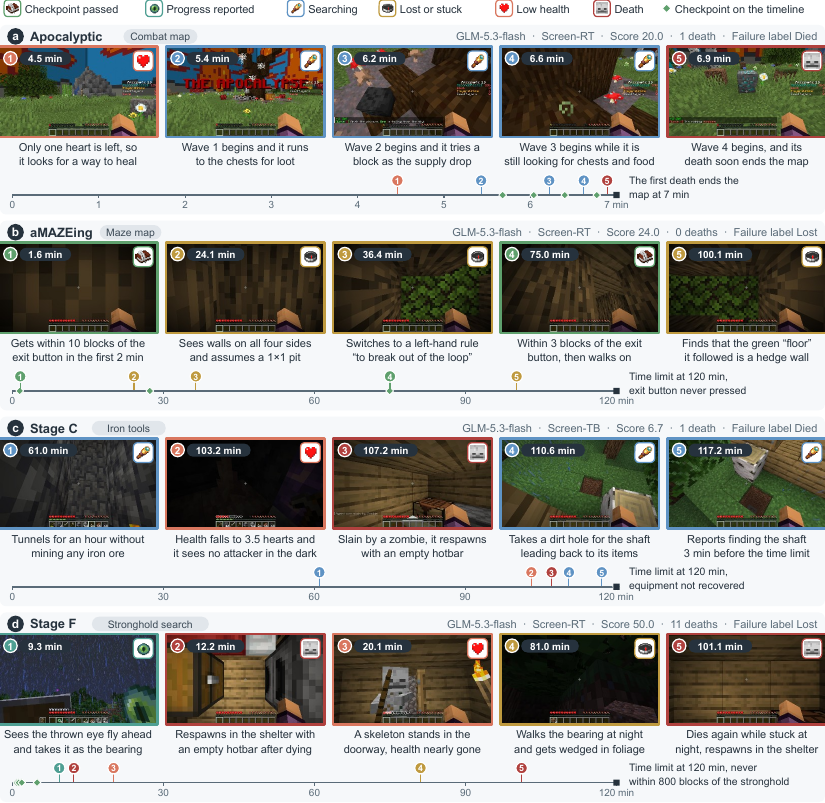}
\caption{Four recorded GLM-5.3-flash runs as screenshot strips. Each strip shows five screenshots that the agent received, in time order, with the wall-clock minute since the start of its session. The icon marks the event type, and each caption paraphrases what the agent wrote right after receiving that screenshot or what the verifier recorded. The timeline under each strip places the numbered screenshots and the passed verifier checkpoints on the wall-clock time of the run. Panels a and b are Lite-map runs under Screen-RT, panel c is the Screen-TB run that loses its equipment underground on stage C, and panel d is the Screen-RT run on stage F. Failure labels follow Table~\ref{tab:failures}.}
\label{fig:case-strips}
\end{figure*}

\paragraph{Community-map cases.}
\label{app:community-failures}
Low community-map scores stem from distinct failure modes, including imprecise low-level control and failure to infer map-specific rules. Two Lite maps illustrate these modes, based on model-assisted trace annotations of the DeepSeek V4.1 Flash and GLM-5.3-flash runs. On the parkour map Easy Parkour!!!, all four Screen-RT runs score zero: the annotated Screen runs repeatedly miss the opening chain-and-lava sequence and never clear the first section, whereas Basic tools earn partial credit for three systems (four-system mean 9.0). On the combat map Apocalyptic, Screen-RT scores range from 5 to 20 (mean 15.0) while the Basic-RT mean is lower at 6.3. In the annotated Screen runs, DeepSeek V4.1 Flash dies early without engaging enemies and GLM-5.3-flash searches for food before dying, as shown in Figure~\ref{fig:case-strips}a, indicating that neither run infers the map's wave-survival rules.

\paragraph{Failure categories.}
\label{app:failure-labels}
Table~\ref{tab:failures} summarizes the primary failure categories across the 448 main-evaluation runs using the Screen interface or Basic tools (four systems, two clocks, and 8 stages or 20 Lite maps, yielding 64 stage runs and 160 Lite-map runs per interface). Solved runs are counted under Full score and excluded from failure labeling. For each remaining run, a labeling agent assigned one primary category from a fixed taxonomy with a one- to two-sentence justification, based on extracted run-record features (score, deaths, termination type and turn count), the final assistant message and key trajectory excerpts. Among Screen runs, navigation loss (Lost) is the most common label (65 of 224 runs), followed by early stopping (55) and death (50), whereas unsolved Basic runs are most frequently attributed to death (64 of the 195 unsolved runs). Figure~\ref{fig:case-strips}b shows a Screen-RT run labeled Lost on the maze map aMAZEing, in which the agent comes within 3 blocks of the exit button but never presses it.

\begin{table}[!htbp]
\centering\small
\caption{Primary failure category of the Screen and Basic runs of the main evaluation, pooled over the two clocks. Full score counts solved runs, and Other covers tool, interface and infrastructure errors.}
\label{tab:failures}
\vspace{2pt}
\setlength{\tabcolsep}{2.8pt}
\begin{tabularx}{\linewidth}{l R R R R >{\hsize=1.25\hsize\raggedleft\arraybackslash}X >{\hsize=.75\hsize\raggedleft\arraybackslash}X R R R}
\toprule
\lchead Group & Died & Stopped & Control & Lost & Knowledge & Slow & Other & Full score & Runs \\
\midrule
\rowcolor{LCBlueFill}Stages, Screen & 10 & 23 & 15 & 14 & 0 & 2 & 0 & 0 & 64 \\
\rowcolor{LCTealFill}Stages, Basic & 19 & 11 & 2 & 4 & 0 & 7 & 3 & 18 & 64 \\
\rowcolor{LCBlueFill}Lite maps, Screen & 40 & 32 & 14 & 51 & 17 & 1 & 5 & 0 & 160 \\
\rowcolor{LCTealFill}Lite maps, Basic & 45 & 32 & 16 & 31 & 14 & 9 & 2 & 11 & 160 \\
\bottomrule
\end{tabularx}
\end{table}

\paragraph{Recovery after setbacks.}
\label{sec:discussion}
Two recorded GLM-5.3-flash trajectories, shown in Figure~\ref{fig:case-strips}c and d, illustrate how a single setback can compound over a long run. In a Screen-TB run, the agent lost its equipment after dying underground and spent the remainder of the run searching for its death location, even with the world paused during reasoning. In a Screen-RT run on stage F, the agent obtained the stronghold bearing but became disoriented in a forest at night, died 11 times, and never came within 800 blocks of the target.

\FloatBarrier

%% file: sections/appendix_dataset.tex
\section{Dataset Analysis}
\label{app:dataset}
\label{sec:dataset}
\begin{figure}[!htbp]
\centering
\begin{subfigure}[b]{0.49\linewidth}
\centering
\includegraphics[width=\linewidth]{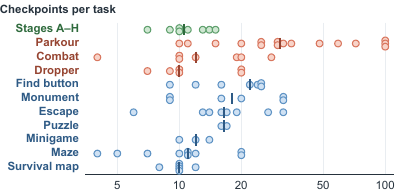}
\caption{Checkpoints per Task}
\end{subfigure}\hfill
\begin{subfigure}[b]{0.49\linewidth}
\centering
\includegraphics[width=\linewidth]{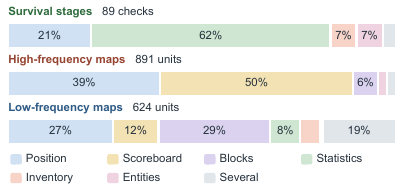}
\caption{What the Checkpoints Read}
\end{subfigure}
\caption{Dataset statistics for the eight survival stages and 70 community maps. (a) Distribution of checkpoints per task on a logarithmic scale, with ticks marking category medians. (b) Distribution of world-state signal types read by verifier checkpoints across survival stages, high-frequency maps and low-frequency maps.}
\label{fig:dataset}
\end{figure}

\paragraph{Checkpoint density.}
Dense checkpoint coverage allows task scores to resolve intermediate progress rather than only binary completion. As shown in Figure~\ref{fig:dataset}a, the eight survival stages contain 7 to 15 checks each (89 in total), and the 70 community maps contain 1,515 weighted units, with a median of 16 per map and a range of 4 to 100. Multi-stage parkour courses are the densest at a median of 31 units, while every category provides multiple intermediate checkpoints. Wall-clock budgets total 196 hours per setting: 16 hours across the eight stages (2 hours each), 170 hours across the 70 community maps (40 at 2 hours and 30 at 3 hours) and 10 hours for the continuous mainline.

\paragraph{Verifier state signals.}
The task families require the verifier to inspect distinct parts of the game's world state, as shown in Figure~\ref{fig:dataset}b. In the survival stages, 62\% of checks read game statistics and advancements, such as crafted items or dimension entry, and 21\% test player position or surroundings. Units on high-frequency maps predominantly read scoreboards or tags set by map logic (50\%) and player position (39\%). Units on low-frequency maps span a broader range of signals: 29\% inspect placed or removed blocks, 27\% positions, 12\% scoreboards and 19\% combine multiple state sources. Evaluating the full benchmark therefore requires all six signal types.

\paragraph{Provenance and properties.}
The community maps come from seven public repositories and at least 42 creators: 28 from Mapcraft, 20 from CurseForge and 11 from one parkour creator's site. For the 26 maps whose source pages report downloads, counts reach 10.6 million with a median of about 60,000. Thirty-six maps target Minecraft 1.21, 30 were built for versions 1.16 to 1.20 and adapted to the benchmark's 1.21.1 environment, and four have no recorded version. The maps emphasize visual perception and control over reading comprehension: among the 61 maps assessed for text reliance, 54 require little text, and six of the other seven are escape or puzzle maps. Ten maps enforce map-specific failure conditions that end scoring early, including six combat arenas.
\FloatBarrier

%% file: sections/appendix_tasks.tex
\section{Minecraft Environment and Task Details}
\label{app:tasks}
\subsection{Task Collections}
We build all tasks in Minecraft Java Edition 1.21.1, where the agent acts from a first-person view through movement, camera control and block or entity interactions. Table~\ref{tab:tasks} summarizes the three task collections, how each task starts, how it is scored and its wall-clock budget.

\begin{table}[!htbp]
\centering\small
\caption{Task collections. Lite and Full give the number of tasks in \benchmark{}-Lite and \benchmark{}-Full.}
\label{tab:tasks}
\label{tab:community-categories}
\vspace{2pt}
\setlength{\tabcolsep}{4pt}
\begin{tabularx}{\linewidth}{l c c >{\raggedright\arraybackslash}X >{\raggedright\arraybackslash}X r}
\toprule
\lchead Collection & Lite & Full & Start and instruction & Checkpoints & Budget \\
\midrule
Survival stages & 8 & 8 & Prepared world state (S0, S\textsubscript{A}--S\textsubscript{G}), stage goal & 89 checks & 2 h each \\
Continuous mainline & 1 & 1 & Empty-handed S0, ``Defeat the Ender Dragon'' & 76 checkpoints & 10 h \\
Community maps & 20 & 70 & Map's own start and goal & 100 points per map (298 weighted units on Lite, 1,515 on Full) & 2 or 3 h \\
\midrule
Total & 29 & 79 & & & \\
\bottomrule
\end{tabularx}
\end{table}

\paragraph{Community-map categories.}
The 70 community maps fall into ten categories. Each category contributes two Lite maps, and the other 50 maps are distributed unevenly: 14 parkour, 4 dropper, 7 maze, 5 find-button, 6 escape, 4 monument, 5 combat, 4 survival and 1 minigame map, with no puzzle map beyond the two Lite maps. Table~\ref{tab:maps} lists every map with its category and creator.

\paragraph{Map curation.}
Candidate maps were screened along six documented dimensions: mechanic coverage (assignment to one of the ten categories), task variation (diversity in route length, difficulty, creator and interaction pattern so that paired Lite maps differ), engine compatibility (verified starting state, commands and interactions in Minecraft 1.21.1), progress signals (native stage counters, scoreboards, in-map checkpoints or inventory targets readable by the verifier), execution evidence (reference runs and recordings used to inspect task behavior) and source metadata (creator, source page, original game version and dated download or view counts when available).

\paragraph{Startup check and scoring examples.}
Startup verification confirms that all 70 maps initialize cleanly in our Minecraft 1.21.1 environment, based on 50 idle startup runs for the additional Full maps and the 20 Lite-map runs. For example, Easy Parkour!!! scores 11 weighted route-position units, Apocalyptic scores wave thresholds while treating the first death as terminal failure, and aMAZEing combines exit-button proximity with transitions across three maze levels.

\FloatBarrier
\subsection{Stage Checks in the Continuous Mainline}
The continuous mainline reuses the checks of the eight survival stages, enabling direct comparison between staged and continuous progression. Of the 89 stage checks, 74 retain identical predicates in the mainline, 2 drop a final-position qualifier and 13 are excluded, yielding the 76 scored checkpoints listed per stage in Table~\ref{tab:stage-map}. The mainline also tracks 18 unscored auxiliary milestones.

\begin{table}[!htbp]
\centering\small
\caption{Stage checks and their treatment in the continuous mainline. The two modified checks remove a final-position qualifier.}
\label{tab:stage-map}
\vspace{2pt}
\begin{tabularx}{\linewidth}{l R R R R}
\toprule
\lchead Stage & Stage checks & Identical & Modified & Excluded \\
\midrule
A: Wood and shelter & 14 & 12 & 0 & 2 \\
B: Stone tools and food & 13 & 12 & 0 & 1 \\
C: Iron tools & 15 & 12 & 0 & 3 \\
D: Diamond mining & 10 & 7 & 1 & 2 \\
E: Nether resources & 11 & 8 & 0 & 3 \\
F: Stronghold search & 10 & 9 & 0 & 1 \\
G: End-portal access & 7 & 6 & 0 & 1 \\
H: Dragon fight & 9 & 8 & 1 & 0 \\
\midrule
Total & 89 & 74 & 2 & 13 \\
\bottomrule
\end{tabularx}
\end{table}

\FloatBarrier

%% file: sections/appendix_gallery.tex
\subsection{Task Gallery}
\label{app:gallery}
\begin{figure}[!htbp]
\centering
\includegraphics[width=\textwidth]{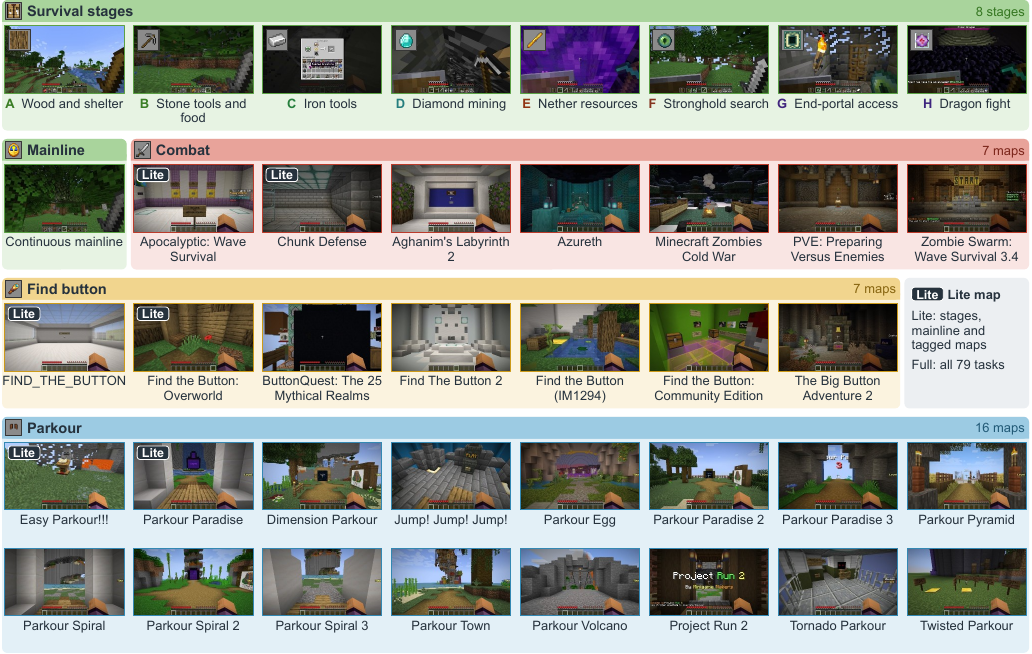}
\caption{Task gallery, part 1: the eight survival stages, the continuous mainline and the Combat, Find button and Parkour maps. Each tile shows a recorded screenshot of one task, each header names a category and its map count, and Lite tags mark the \benchmark{}-Lite maps.}
\label{fig:gallery-1}
\end{figure}

\begin{figure}[!htbp]
\centering
\includegraphics[width=\textwidth]{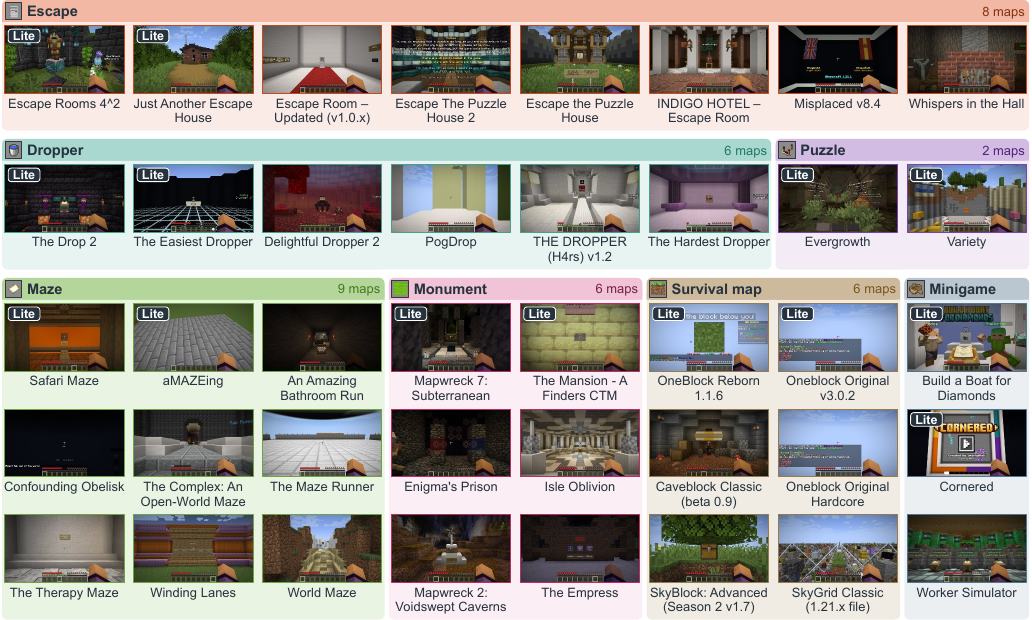}
\caption{Task gallery, part 2: the Escape, Dropper, Puzzle, Maze, Monument, Survival map and Minigame maps, drawn as in Figure~\ref{fig:gallery-1}.}
\label{fig:gallery-2}
\end{figure}

Figure~\ref{fig:gallery-1} and Figure~\ref{fig:gallery-2} show one screenshot for each of the 79 tasks of \benchmark{}-Full: the eight survival stages, the continuous mainline and the 70 community maps in their ten categories, under the names used in Table~\ref{tab:maps}. The stage and mainline screenshots are frames from recorded Claude Opus 5.5 runs of those tasks. Each map screenshot is the first panel of the report made for that map during curation, except for Easy Parkour!!!, whose screenshot comes from a recorded run. The screenshots illustrate the scenes an agent faces and carry no information about task outcomes. The icons are item and block textures from Minecraft 1.21.1, the game version of the benchmark.

\FloatBarrier

%% file: sections/appendix_tools.tex
\section{Interaction Tools}
\label{app:tools}
All tools are served by the game container under identical task specifications, instructions, budgets and verifiers.

\subsection{Screen, Screen + State and Basic Tools}
The Screen interface provides two primitives: \texttt{look}, which returns a $960\times540$ screenshot, and \texttt{act}, which holds keys, rotates the camera or clicks for up to 400 game ticks before returning a new screenshot. Screen + State appends the player's position, orientation, health, hunger, inventory and targeted block as text. The Basic tools are the 26 tools of Table~\ref{tab:tools-basic}: 18 local action and inspection primitives, and 8 line-of-sight observation and spatial-memory tools. Basic actions run sequentially without background behaviors between calls.

\clearpage
{\footnotesize
\setlength{\tabcolsep}{4pt}
\begin{longtable}{>{\raggedright\arraybackslash}p{0.21\linewidth} >{\raggedright\arraybackslash}p{0.44\linewidth} >{\raggedright\arraybackslash}p{0.30\linewidth}}
\caption{The 26 Basic tools. The Macro tools include the observation and memory tools under the same names, without the line-of-sight occlusion filter.}\label{tab:tools-basic}\label{tab:tools-obs}\\
\toprule
\rowcolor{LCTealSoft}Tool & Function & Limit \\
\midrule
\endfirsthead
\toprule
\rowcolor{LCTealSoft}Tool & Function & Limit \\
\midrule
\endhead
\multicolumn{3}{@{}l}{\textit{Action and inspection}} \\
\texttt{observe} & Summary of own state, unoccluded entities and nearby visible blocks & Compatibility summary \\
\texttt{inventory} & All inventory slots & None \\
\texttt{recipe} & Recipe lookup: ingredients, output count, crafting-table need & None \\
\texttt{go\_to} & Walk to a point & Within 64 blocks, no digging or bridging, at most 20 s \\
\texttt{move} & Hold movement keys, jump or sprint for $N$ ticks & At most 400 ticks \\
\texttt{look\_at} & Face a point & None \\
\texttt{mine\_block} & Mine one block, choosing the best tool & Within 4.5 blocks and visible \\
\texttt{place\_block} & Place one block in an empty cell & Within 4.5 blocks, next to a solid block \\
\texttt{attack} & Hit one entity several times & Within 3.5 blocks and visible, with cooldown between hits \\
\texttt{use\_block} & Right-click a block: door, lever, button, bed & Within 4.5 blocks and visible \\
\texttt{use\_entity} & Right-click an entity to interact, mount or dismount & Within 3.5 blocks and visible \\
\texttt{use\_item} & Use the held item: eat, drink, draw a bow, throw, use a bucket & None \\
\texttt{equip} & Equip to hand, off-hand or armor slot & None \\
\texttt{drop} & Drop items & None \\
\texttt{craft} & Craft one recipe & $3\times3$ recipes need a crafting table within 4.5 blocks \\
\texttt{chest} & Open a container to store, take or list items & Within 4.5 blocks \\
\texttt{furnace} & Add input and fuel, take output, read progress & Within 4.5 blocks \\
\texttt{wait} & Wait & At most 400 ticks \\
\midrule
\multicolumn{3}{@{}l}{\textit{Observation and memory}} \\
\texttt{view\_status} & Health, hunger, position, equipment, and unoccluded mobs, dragon state and crystals & None \\
\texttt{view\_blocks} & Search loaded chunks for unoccluded blocks or query a box or cell list (occluded cells return as unobserved) & None \\
\texttt{view\_frontier} & Biome and surface-water map of remembered chunks and unexplored frontier & None \\
\texttt{query\_bot\_data} & Read-only SQL over own state, inventory, known map, opened containers and notes & None \\
\texttt{read\_recent\_events} & Recent events: player messages, deaths, dimension changes, equipment & None \\
\texttt{view\_crafting\_\allowbreak requirements} & Recursive recipe and missing materials, without crafting & None \\
\texttt{note\_save}, \texttt{note\_read} & Save and read persistent notes & None \\
\bottomrule
\end{longtable}
}

\FloatBarrier
\subsection{Macro Tools and Clock Scheduling}
The Macro tools, adapted from mine-ai-mcp, are the 48 tools in Table~\ref{tab:tools-macro} under the real-time clock, pairing loaded-chunk observations, including occluded blocks, with multi-step routines and background survival behaviors, as contrasted with Basic tools in Table~\ref{tab:tools-compare}. Under the turn-based clock, the world advances at 20 ticks per second only while an action executes and pauses as soon as the tool returns, so reasoning and read-only queries consume zero game ticks. In Macro-TB, long actions execute synchronously and asynchronous wait tools are omitted.

\begin{table}[!htbp]
\centering\footnotesize
\caption{Macro tools under the real-time clock, by category.}
\label{tab:tools-macro}
\vspace{2pt}
\setlength{\tabcolsep}{4pt}
\begin{tabularx}{\linewidth}{l >{\raggedright\arraybackslash}X}
\toprule
\rowcolor{LCSandSoft}Category & Tools \\
\midrule
Perception and memory & \texttt{view\_status}, \texttt{view\_blocks}, \texttt{view\_frontier}, \texttt{view\_crafting\_\allowbreak requirements}, \texttt{query\_bot\_data}, \texttt{read\_recent\_events}, \texttt{note\_save}, \texttt{note\_read} \\
Movement and exploration & \texttt{navigate}, \texttt{explore\_frontier}, \texttt{locate\_stronghold}, \texttt{enter\_nether\_portal}, \texttt{enter\_end\_portal}, \texttt{activate\_portal} \\
Collection and building & \texttt{collect\_block}, \texttt{collect\_mob\_drop}, \texttt{pick\_up\_items}, \texttt{place\_block}, \texttt{build\_structure}, \texttt{use\_bucket} \\
Items & \texttt{craft\_item}, \texttt{smelt\_item}, \texttt{use\_container}, \texttt{equip}, \texttt{eat\_food}, \texttt{drop\_item}, \texttt{barter} \\
Local actions & \texttt{use\_block}, \texttt{use\_entity}, \texttt{use\_item}, \texttt{mine\_block}, \texttt{attack}, \texttt{move}, \texttt{look\_at}, \texttt{chest}, \texttt{furnace}, \texttt{wait}, \texttt{list\_entities} \\
Combat and the dragon & \texttt{destroy\_end\_crystal}, \texttt{attack\_dragon\_perch}, \texttt{prepare\_dragon\_perch}, \texttt{shoot\_dragon} \\
Other & \texttt{set\_survival\_policy}, \texttt{sleep}, \texttt{raw\_action}, \texttt{send\_message}, \texttt{wait\_for\_action}, \texttt{cancel\_foreground\_action} \\
\bottomrule
\end{tabularx}
\end{table}

\begin{table}[!htbp]
\centering\footnotesize
\caption{Capabilities of the Basic and Macro tools.}
\label{tab:tools-compare}
\vspace{2pt}
\setlength{\tabcolsep}{4pt}
\begin{tabularx}{\linewidth}{l >{\raggedright\arraybackslash}X >{\raggedright\arraybackslash}X}
\toprule
\lchead & \cellcolor{LCTealSoft}Basic & \cellcolor{LCSandSoft}Macro \\
\midrule
Blocks & Unoccluded blocks in loaded chunks & All loaded chunks, including buried ores \\
Entities & Unoccluded entities & All loaded entities \\
Movement & \texttt{go\_to} within 64 blocks (no digging or bridging), \texttt{move} holds keys & \texttt{navigate} long-range pathfinding with digging and bridging, \texttt{explore\_frontier} \\
Collection & \texttt{mine\_block} mines one reachable, visible block & \texttt{collect\_block} finds, reaches, mines and picks up $N$ blocks \\
Crafting & \texttt{craft} performs one recipe & \texttt{craft\_item} crafts recursively, \texttt{smelt\_item} waits for smelting \\
Combat & \texttt{attack} hits one visible entity & Background combat policy and dedicated dragon-fight tools \\
Background behavior & None: the player stands still between calls & Eating, evading, fall and fire protection \\
Long actions & Sequential, at most 400 ticks or 20 s per call & Asynchronous under the real-time clock, with progress queries \\
\bottomrule
\end{tabularx}
\end{table}

\FloatBarrier

%% file: sections/appendix_limitations.tex
\section{Limitations and Ethics Statement}
\label{app:limitations}
\paragraph{Limitations.}
Our evaluation covers four agent systems, and extending it to additional models and agent harnesses would clarify the generality of these findings. All systems are evaluated at a single medium reasoning effort, leaving the impact of higher or lower effort budgets to future study. Finally, while the 70 community maps span ten mechanic categories, incorporating additional maps would further broaden task diversity and benchmark scale.

%% file: sections/statements.tex
\paragraph{Ethics.}
\label{sec:ethics}
\benchmark{} builds upon community-authored Minecraft maps and prior research. We respect the rights of the map creators and of Mojang Studios: all community maps are used strictly for non-commercial research, with known creators credited in Table~\ref{tab:maps}, and all foundational prior works are cited throughout the paper. This work is not affiliated with or endorsed by Mojang Studios or Microsoft.
\FloatBarrier

%% file: sections/appendix_maps.tex
\clearpage
\section{Community Maps}
\label{app:maps}
\fancyhead[R]{\footnotesize\sffamily\color{LCMuted}\thesection\hspace{.6em}Community Maps}
Table~\ref{tab:maps} lists the 70 community maps, 20 of which are Lite maps. Source pages are recorded in the benchmark's map metadata, and full starting states, instructions and scoring predicates are provided in the benchmark task files.

{\footnotesize
\setlength{\tabcolsep}{3pt}\setlength{\extrarowheight}{.15ex}
\begin{longtable}{r >{\raggedright\arraybackslash}p{0.10\linewidth} >{\raggedright\arraybackslash}p{0.165\linewidth} >{\raggedright\arraybackslash}p{0.125\linewidth} c >{\raggedright\arraybackslash}p{0.488\linewidth}}
\caption{The 70 community maps, sorted by category with Lite maps first. The Set column distinguishes maps included in \benchmark{}-Lite (Lite) from those unique to \benchmark{}-Full (Full). Descriptions reproduce the creators' official text where reachable ($^\ddagger$ marks a short excerpt) and otherwise give the project's task brief ($^\dagger$).}\label{tab:maps}\\
\toprule
\lchead \# & Category & Map & Author & Set & Description \\
\midrule
\endfirsthead
\toprule
\lchead \# & Category & Map & Author & Set & Description \\
\midrule
\endhead
\bottomrule
\endfoot
1 & Parkour & Easy Parkour!!! & Smaximum\_yt & Lite & Complete the tower parkour and final passage through the iron, diamond and obsidian sections to reach the ending signs, using chest supplies along the way. Death returns you to the start while retaining earned points. $^\dagger$ \\
2 & Parkour & Parkour Paradise & Hielke & Lite & Parkour Paradise is a parkour map which contains 100 little levels, and your goal is to beat them all! The first levels will be easy, but the parkour gets harder the further you get. Are you skilled enough to reach the end? \\
3 & Parkour & Dimension Parkour & Hielke & Full & Dimension Parkour takes you on an amazing parkour journey through all three Minecraft dimensions. Can you complete the parkour and defeat the Ender Dragon at the end? Come find out! \\
4 & Parkour & Jump! Jump! Jump! & JUDEH & Full & A vertical parkour map with three 350-block-high levels selected from a central lobby. Each level grants jump boost and speed without intermediate checkpoints, and pressing the top button clears the level and returns to the lobby. $^\dagger$ \\
5 & Parkour & Parkour Egg & Hielke & Full & Climb the colorful Parkour Egg to the top! There are 10 Easter eggs hidden in the map, can you find them all? $^\ddagger$ \\
6 & Parkour & Parkour Paradise 2 & Hielke & Full & Parkour Paradise is back! Parkour Paradise 2 offers 100 new and exciting levels to beat! With more than double the size of the first one, there's more parkour for you to enjoy! Are you up for the challenge? \\
7 & Parkour & Parkour Paradise 3 & Hielke & Full & Parkour Paradise 3 is a MASSIVE parkour map containing 100 fun and exciting levels to beat! Play alone or with friends! \\
8 & Parkour & Parkour Pyramid & Hielke & Full & Parkour Pyramid is an epic parkour adventure that will keep you entertained for hours! The goal of the map is to reach the top of the pyramid. You start all the way at the bottom. $^\ddagger$ \\
9 & Parkour & Parkour Spiral & Hielke & Full & In Parkour Spiral your goal is to reach the top of the enormous spiral tower! Along the way you have to beat all kind of parkour challenges. Are you skilled enough to reach the end? \\
10 & Parkour & Parkour Spiral 2 & Hielke & Full & In Parkour Spiral 2 your goal is to climb the Spiral from the Nether to the top! Along the way you have to beat a lot of new parkour challenges. \\
11 & Parkour & Parkour Spiral 3 & Hielke & Full & Parkour Spiral 3 offers a GIGANTIC new spiral tower for you to climb! Are you skilled enough to reach the top? Come find out! \\
12 & Parkour & Parkour Town & Hielke & Full & Climb the Parkour Town with its countless pathways to the top! Explore the town, find the hidden coins, and buy powerups to unlock new pathways to improve your time! How fast can you get to the top? \\
13 & Parkour & Parkour Volcano & Hielke & Full & Welcome to Parkour Volcano, where your goal is to reach the top of an active volcano! Each section of the volcano brings new tests of your agility and speed. Can you make it out of the volcano safely? \\
14 & Parkour & Project Run 2 & Minigame Makers (100+ builders) & Full & A multi-author parkour world of connected square plots with branching routes and puzzle mechanics. Open the main gate from the lobby and traverse plots in sequence toward an ending plot. $^\dagger$ \\
15 & Parkour & Tornado Parkour & Jerries & Full & Climb a tornado by parkour to rescue the mayor at the top. Touch glowing markers in sequence to update your respawn checkpoint while periodic crosswinds push you sideways. $^\dagger$ \\
16 & Parkour & Twisted Parkour & HavyFI & Full & Climb a dark twisted tower from the ground to the top platform using blocks that grant levitation, slow falling or teleportation and checkpoints that update your respawn point. $^\dagger$ \\
17 & Dropper & The Drop 2 & Chocobodo & Lite & Steer through tall vertical shafts to land in the water at the bottom without natural health regeneration, then press each shaft's bottom button to advance to the next level. $^\dagger$ \\
18 & Dropper & The Easiest Dropper & DrJones4 & Lite & Glide with an elytra from a sky platform to ground teleporters for nine dropper levels, clearing each shaft by landing in water or pressing the terminal button. $^\dagger$ \\
19 & Dropper & Delightful Dropper 2 & EdanEA, jwicker6 & Full & Have you ever played a dropper map whose difficulty is more infuriating than challenging?\ldots{} $^\ddagger$ \\
20 & Dropper & PogDrop & TheP8 / Vankkeri & Full & A 20-level dropper map across four themed wings (Nether, The End, Nature and Underground). Survive each fall onto water, slime, cobwebs or vines and press the bottom button to unlock the next level in that wing. $^\dagger$ \\
21 & Dropper & THE DROPPER (H4rs) v1.2 & H4rs & Full & A ten-level dropper map with three hearts of health. Survive each fall into a bottom room and locate its hidden button using chat hints to unlock the next corridor door. $^\dagger$ \\
22 & Dropper & The Hardest Dropper & DrJones4 & Full & A half-heart dropper map where any fall damage is fatal. Land in the bottom water of each shaft to return to the lobby and unlock the next level's corridor door. $^\dagger$ \\
23 & Maze & Safari Maze & Das\_Skou & Lite & Search a large hedge maze to collect five levers from the House, Well, Horse Stable, Windmill and Farm, then place and activate all five on iron blocks in the underground puzzle room. $^\dagger$ \\
24 & Maze & aMAZEing & Shumanman & Lite & Traverse three sequential hedge mazes (23$\times$23, 29$\times$29 and 39$\times$39) by finding and pressing the hidden blackstone exit button on the wall of each maze. $^\dagger$ \\
25 & Maze & An Amazing Bathroom Run & TheZaius & Full & Navigate a dark corridor maze of tight passages to find the room with a toilet and use it to reach the final room. $^\dagger$ \\
26 & Maze & Confounding Obelisk & DeuxiemeCarlin & Full & A vertical tower of stacked 80$\times$80 themed floor mazes where stepping into the opposite exit booth teleports you to the next floor. $^\dagger$ \\
27 & Maze & The Complex: An Open-World Maze & DeuxiemeCarlin & Full & Welcome to The Complex: An Open-World Maze!!! As the name suggests, this map is an open-world maze map- and it's pretty complex, too! \ldots{} $^\ddagger$ \\
28 & Maze & The Maze Runner & Samster99 \& David1220 & Full & Starting in a central clearing surrounded by 100-block-high walls, locate and pass through the maze exit without breaking or climbing the walls. $^\dagger$ \\
29 & Maze & The Therapy Maze & Simon\v{C}eli \& MrYellow & Full & Navigate a narrow lit maze to find the exit sign and press its button while ignoring the locked decoy door beside it. $^\dagger$ \\
30 & Maze & Winding Lanes & DeuxiemeCarlin & Full & A hub maze connecting twelve themed mazes, each completed by pressing a polished blackstone button at its end. $^\dagger$ \\
31 & Maze & World Maze & AsianPlaneBoi & Full & Traverse a long multi-theme maze floating over the void while avoiding floor holes and a reset button partway through. $^\dagger$ \\
32 & Find button & FIND\_THE\_BUTTON & leandrov & Lite & 'FIND THE BUTTON' is a map, as the name suggests, where you can find buttons spread across 10 maps. \ldots{} $^\ddagger$ \\
33 & Find button & Find the Button: Overworld & Mento & Lite & Search a sequence of small themed rooms to find and press the single hidden button that teleports you to the next level. $^\dagger$ \\
34 & Find button & ButtonQuest: The 25 Mythical Realms & Yuliun, GHJURD \& Xmmaud & Full & Search 25 themed realms to find and press the target button type specified in the goal while ignoring decoy buttons. $^\dagger$ \\
35 & Find button & Find The Button 2 & d\_de\_diego\_d & Full & In this Minecraft map, your goal is to find a hidden button in each room to advance. \ldots{} It has 25 levels set in each dimension of Minecraft $^\ddagger$ \\
36 & Find button & Find the Button (IM1294) & FireMaster1294 & Full & This is a take on the classic Find the Button featuring builds designed around many 1.18.2 structures and biomes\ldots{} Now available for 1.20.5! Please leave a comment if you find any bugs as Mojang changed a lot of things surrounding data tagging. $^\ddagger$ \\
37 & Find button & Find the Button: Community Edition & Drag0n\_Tamer & Full & Search community-built levels (houses, castles, a school and a mineshaft) to locate the real button among trap buttons that damage, blind or teleport you. $^\dagger$ \\
38 & Find button & The Big Button Adventure 2 & Milu012345 & Full & Welcome to one of the biggest Find The Button maps ever! \ldots{} $^\ddagger$ \\
39 & Escape & Escape Rooms 4\textasciicircum{}2 & Fireless007 & Lite & 4\textasciicircum{}2 Escape Rooms is a map that will challenge and question your skills and your knowledge of Minecraft. With a total of 16 challenges, you will have the time to question yourself. There are different types of challenges :\ldots{} $^\ddagger$ \\
40 & Escape & Just Another Escape House & Suhn Jhi & Lite & Solve visual clues across item frames, flower pots, clocks and block patterns to unlock doors and escape a locked brick house. $^\dagger$ \\
41 & Escape & Escape Room -- Updated (v1.0.x) & Ricciodilego & Full & Challenge yourself! IMPORTANT: For having a better experience, play this map using the related texture. This map has 5 levels, each one will test your skills. \\
42 & Escape & Escape The Puzzle House 2 & Andersbahl1234 & Full & Explore a multi-floor mansion, read clue books and picture codes to unlock rooms and escape while collecting up to ten hidden ``Point'' items. $^\dagger$ \\
43 & Escape & Escape the Puzzle House & Andersbahl1234 & Full & Enter a mansion from the front yard, solve chained book and key puzzles across its floors to escape and collect hidden ``Point'' items. $^\dagger$ \\
44 & Escape & INDIGO HOTEL -- Escape Room & henrico & Full & This is an adventure map focused on escape rooms, where you must escape from 100 rooms with puzzles, mazes, monsters\ldots{} This map was made to Minecraft: Java Edition - version 1.20.1 $^\ddagger$ \\
45 & Escape & Misplaced v8.4 & InfiniteXgg & Full & After nocliping through the floor you find yourself in a strange place out of our reality, seems like nobody is nearby\ldots{} This is a 1.21.6+ minecraft map that counts with it's own resource pack and datapack that complements each other $^\ddagger$ \\
46 & Escape & Whispers in the Hall & That Dude & Full & Infiltrate the dark MythRelic Museum at night, read security-room books, restore power while avoiding moving statues and escape through the ventilation shafts. $^\dagger$ \\
47 & Puzzle & Evergrowth & Gamemode 4 & Lite & A block pushing puzzle map for Minecraft Java. Time to get pushing.. Evergrowth has a simple idea at its heart... pushing blocks: But things can slowly get complicated as we add more mechanics... like ice, teleporters or (shown here) magnets! $^\ddagger$ \\
48 & Puzzle & Variety & FatalVortex & Lite & Don't know what game mode to play? Well, play everything in Variety. A map featuring eight different types of Minecraft game modes including Maze, Parkour, Find the Button, Puzzle, Dropper\ldots{} This is Variety. A map where you can play eight different game modes at any time you want. $^\ddagger$ \\
49 & Monument & Mapwreck 7: Subterranean & CTM Community & Lite & An underground Complete-the-Monument map of 16 multi-author areas, each cleared by obtaining its wool-named objective item. $^\dagger$ \\
50 & Monument & The Mansion - A Finders CTM & user\_\allowbreak{}zysb6jpz\allowbreak{}jav9in4r & Lite & The Mansion is an immersive Complete the Monument adventure, set in an old, sprawling mansion filled with secrets\ldots{} $^\ddagger$ \\
51 & Monument & Enigma's Prison & Link2012 & Full & Hardcore map enjoyers rejoice! \ldots{} $^\ddagger$ \\
52 & Monument & Isle Oblivion & DINOCREATOR \& Phoraxe & Full & An island Complete-the-Monument survival map where you explore themed areas to recover 16 colored wools and place them on the Serenity Monument. $^\dagger$ \\
53 & Monument & Mapwreck 2: Voidswept Caverns & CTM Community & Full & A subterranean Complete-the-Monument map where you recover 16 colored wools across multi-author cavern areas and place them in the monument hall. $^\dagger$ \\
54 & Monument & The Empress & q2kro \& ImSheol & Full & Starting from a shipwreck, clear the areas of a castle empire to recover eight colored wools and place them on the monument of eight knights. $^\dagger$ \\
55 & Combat & Apocalyptic: Wave Survival & kingzustin & Lite & A hard survival minigame about the end of the world! \ldots{} This is Minecraft, so you can build, break, craft, smelt, and everything when in-game! $^\ddagger$ \\
56 & Combat & Chunk Defense & snakemasterepic & Lite & Single-death wave survival on a 16$\times$16 chunk combining daytime resource gathering and defense building with nighttime monster waves and between-round shop upgrades. $^\dagger$ \\
57 & Combat & Aghanim's Labyrinth 2 & MrKheese & Full & A roguelike combat labyrinth where you clear monster rooms across sequential floors, trade with shop villagers for spells and gear, and consume floor shards to advance. $^\dagger$ \\
58 & Combat & Azureth & Belven & Full & A five-wave underground warped-forest combat arena where each wave requires defeating a target number of spawned monsters. $^\dagger$ \\
59 & Combat & Minecraft Zombies Cold War & Dulmari\_Games & Full & Single-life endless zombie-wave survival where rotten flesh drops unlock doors and equipment upgrades across the arena. $^\dagger$ \\
60 & Combat & PVE: Preparing Versus Enemies & Joepstin & Full & Single-life prepare-and-fight arena cycling every 80 seconds between castle shop upgrades and monster waves, culminating in the Strategy Master boss. $^\dagger$ \\
61 & Combat & Zombie Swarm: Wave Survival 3.4 & kingzustin & Full & Defend the world as you fight off massive swarms of zombies! \ldots{} there get to be as many as 200+ zombies all at once! $^\ddagger$ \\
62 & Survival map & OneBlock Reborn 1.1.6 & Gamerz\_Riot & Lite & Oneblock Reborn is a Adventure map, it is continuation of IJAMinecraft's OneBlock Map. $^\ddagger$ \\
63 & Survival map & Oneblock Original v3.0.2 & Crepi (Crimson Creations) & Lite & Expand a floating sky platform from a single regenerating block that cycles through progressive resource, chest and mob phases above the void. $^\dagger$ \\
64 & Survival map & Caveblock Classic (beta 0.9) & Crepi & Full & Welcome to the underground madness! Build your underground base, find camps and save the villagers! Minecraft survival map inspired by stoneblock modpack. $^\ddagger$ \\
65 & Survival map & Oneblock Original Hardcore & Crepi (Crimson Creations) & Full & Hardcore single-life variant of Oneblock Original where mining the central regenerating block progresses through phases and the first death ends the run. $^\dagger$ \\
66 & Survival map & SkyBlock: Advanced (Season 2 v1.7) & darkfated & Full & The map consists of many flying islands located in the distance. And by appearing on the map, your main goal is to survive. \\
67 & Survival map & SkyGrid Classic (1.21.x file) & noxgame1230 & Full & This is a skygrid-type map. \ldots{} the map does not require mods. $^\ddagger$ \\
68 & Minigame & Build a Boat for Diamonds & TheWorfer27 & Lite & In the depths of the Underworld, the mighty King Hades laughs manically, knowing that he has trapped the world in eternal winter. The only way to reach him is to engineer a boat to sail the River Styx, but it won't be easy. \\
69 & Minigame & Cornered & TheWorfer27 / Smarty & Lite & Cornered. Cornered? Cornered! Inspired by the board game Blokus, the one rule is that your pieces can only touch on their corners \ldots{} $^\ddagger$ \\
70 & Minigame & Worker Simulator & TEHEK & Full & An economy simulator where you start as a harbor loader carrying boxes from a truck to delivery points to earn money for city jobs. $^\dagger$ \\
\end{longtable}
}
\FloatBarrier